\documentclass{article}

\usepackage[preprint,nonatbib]{neurips_2023}

\usepackage[utf8]{inputenc} %
\usepackage[T1]{fontenc}    %
\usepackage{hyperref}       %
\usepackage{url}            %
\usepackage{booktabs}       %
\usepackage{amsfonts}       %
\usepackage{nicefrac}       %
\usepackage{microtype}      %
\usepackage{xcolor}         %
\usepackage{subfigure}
\usepackage{graphicx}
\usepackage{multirow}
\usepackage{array}
\usepackage{amsmath}
\usepackage{amssymb}
 \usepackage{colortbl}
\usepackage{acronym}

\acrodef{LLM}{Large Language-Model}
\acrodef{COCOCap}{COCO Captions}
\acrodef{REG}{Referring Expressions Generation}
\acrodef{GCC}{Google Conceptual Captions}

\newcommand\disclip[1]{DisCLIP}
\DeclareMathOperator*{\argmax}{argmax}

\title{DisCLIP: Open-Vocabulary Referring Expression Generation}

\author{%
  Lior Bracha$^*$\\
  Bar-Ilan University, Israel \\
  \texttt{brachal@biu.ac.il} \\
  \And
  Eitan Shaar\thanks{Equal Contribution}\\
  Bar-Ilan University, Israel \\
  \texttt{shaarei@biu.ac.il} \\
  \And
  Aviv Shamsian \\
  Bar-Ilan University, Israel \\
  \texttt{aviv.shamsian@live.biu.ac.il} \\
  \And
  Ethan Fetaya \\
  Bar-Ilan University, Israel \\
  \texttt{ethan.fetaya@biu.ac.il} \\
  \And
  Gal Chechik \\
  Bar-Ilan University, Israel \\ NVIDIA, Israel \\
  \texttt{gal.chechik@biu.ac.il} \\
}

\begin{document}

\maketitle

\begin{abstract}
  
  Referring Expressions Generation (REG) aims to produce textual descriptions that unambiguously identifies specific objects within a visual scene. Traditionally, this has been achieved through supervised learning methods, which perform well on specific data distributions but often struggle to generalize to new images and concepts. To address this issue, we present a novel approach for REG, named \textit{DisCLIP}, short for discriminative CLIP. We build on CLIP, a large-scale visual-semantic model, to guide an LLM to generate a contextual description of a target concept in an image while avoiding other distracting concepts.
  Notably, this optimization happens at inference time and does not require additional training or tuning of learned parameters. We measure the quality of the generated text by evaluating the capability of a receiver model to accurately identify the described object within the scene. To achieve this, we use a frozen zero-shot comprehension module as a critique of our generated referring expressions. We evaluate DisCLIP on multiple referring expression benchmarks through human evaluation and show that it significantly outperforms previous methods on out-of-domain datasets. Our results highlight the potential of using pre-trained visual-semantic models for generating high-quality contextual descriptions.
  
\end{abstract}

\begin{figure}[hb]
    \centering
    \includegraphics[width=\linewidth, trim={0 0 0 0}, clip]{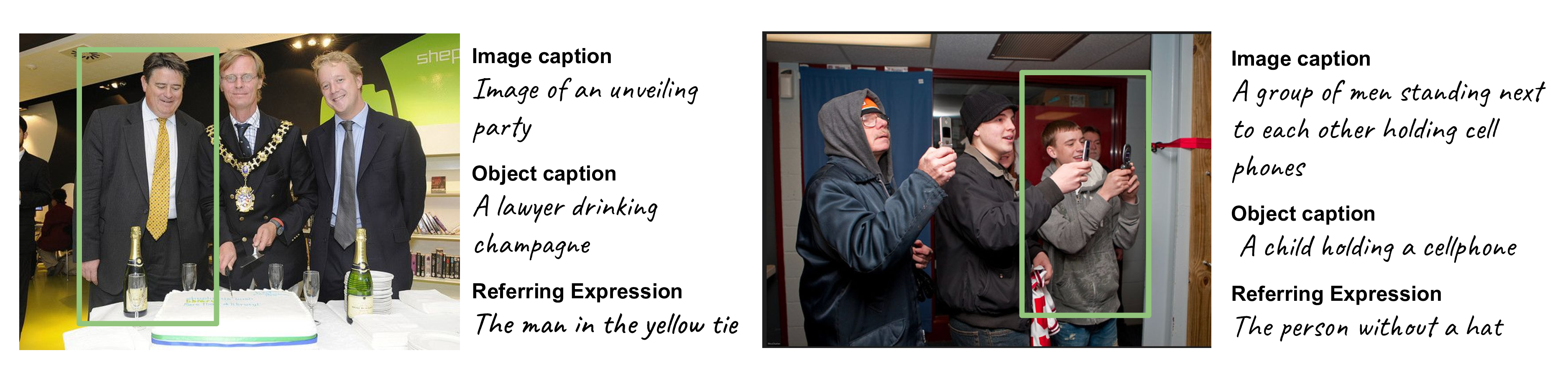} 
    \caption{\small{
    \textbf{Referring Expressions Generation }(REG) aims to generate textual descriptions that clearly identify an object in a given scene, ignoring similar distractors. REG is harder than object (dense) captioning because it must take into account the context of other objects. For instance, a REG model must be capable of identifying unique features such as the color of a tie (on the left) or general descriptions such as "\textit{the man without the hat}" (right). The same object can have multiple distinct descriptions based on the context.}}
    \label{fig:key}
\end{figure} 

\section{Introduction} \label{sec:intro}

Referring expressions (REs) are a key component of language communication. They allow people to refer to one  entity in a complex visual scene, in an unambiguous way. Comprehending and generating REs is essential for embedded agents that need to communicate with people about their environment. For instance, an autonomous vehicle may inquire a passenger about their preferences - ``Should I park in the nearest spot or the shaded one?", or a robot assistant may wish to clarify an instruction: ``Which chair should I get you: the black one or the white one?". 

Significant effort has been devoted to RE \textit{comprehension}, namely, 
training agents to understand referring expressions generated in natural language by people. 
The current paper focuses on a complementary task: \ac{REG}, namely, training agents to refer to entities in a visual scene using natural language. RE generation and comprehension are complementary; they can be viewed as played by two communicating players \cite{andreas2016reasoning,kazemzadeh2014referitgame,andreas-klein-2016-reasoning,vedantam2017context}.
First, a \textit{speaker} observes a scene that contains multiple objects and generates language that refers to a specific target object. Then, a second player, a \textit{listener}, interprets the RE in the context of the same visual scene 
and selects the entity that is referred to. In this communication-as-a-game setup, the two players are cooperative and have a common objective, the speaker wishes to generate REs that are easily interpretable by the listener \cite{vered2019joint,luo2018discriminability}.

\ac{REG} generated by agents should satisfy two key properties: (1) discriminative - using attributes to point clearly to a unique object in the scene, and (2) intelligible - producing language that can be easily understood by people. Recent advances in NLP using web-scale corpora have been very successful in generating natural language, but datasets available for referring expressions are significantly smaller. As a result, current visual REG models are limited and do not transfer well to images outside the narrow domain they were trained on. 
In contrast,

Visual-Linguistic (VL) models, such as CLIP \cite{radford2021learning} and LLM, were trained on large text corpora, encompassing a wide variety of expressions. Therefore, they provide a more versatile and general-purpose framework for REG. In addition, the vast scale of foundation models enables them to generalize effectively to new data, even in zero-shot scenarios. 

Building on these advances we propose an approach that builds on LLMs and large VL models.
Our approach is based on two key components. First, we use a pre-trained CLIP as a listener to evaluate how well a text phrase corresponds to an object in a given scene. Second, we introduce a method for using CLIP in a discriminative manner across localized boxes. This optimization is guiding the text generation of an LLM, at inference time. As such, it does not require any further training of learned parameters.
Importantly, we avoid training the listener and speaker models jointly, because such training can lead to a ``runaway'' drift into a specialized language that is less natural for human interpretation. Furthermore, CLIP was trained on a large text corpora, which allows for an open-vocabulary generation, that effectively generalize to new vision and language domains. 

Referring expressions have been traditionally separated into two types: relational (``the person on the left") and attribute-based (``the person with the hat"). This paper focuses on attribute-based REG because current VL models represent attributes much better than spatial relations. 

This paper makes the following contributions. \textbf{(1)} First, we introduce the first approach to open-vocabulary visual referring expression generation named \textit{DisCLIP} for discriminative CLIP. It allows generalizing to new data distributions and concepts, making it more versatile and adaptable to various applications. Notably, these results are achieved without any additional training or fine-tuning required. \textbf{(2)} We put forward a method that utilizes foundation models trained on image-level descriptions, for the generation of \textit{contextual descriptions}. Such descriptions are costly to curate, and are rarely available, even in large VL corpora. \textbf{(3)} We show through extensive experiments that DisCLIP outperforms supervised methods on out-of-domain datasets in varying learning setups. 
Importantly, our model is producing descriptions that are more natural and accurate, according to human raters.

\section{Related work} \label{sec:lit}

REG task is often considered a proxy of pragmatic reasoning and naturally falls under the paradigm of a dialog. Effective communication and  contextual language are further explored in the Rational Speech Act (RSA) framework \cite{frank2012predicting}. Accordingly, a common architecture is a speaker and a listener, performing complementary tasks: REG and REC. A broad class of REG methods 
\cite{luo2017comprehension,liu2020attribute,mao2016generation,hu2016natural,yu2017joint} rely on joint optimization of the speaker, and the listener. The risk in such a pipeline is creating a ``secret'' language  \cite{vered2019joint} the speaker-listener architecture tends to overfit and struggle to generalize across domains. In contrast, our method does not require any training and entirely depends on inference time decoding. Other notable works in that field include \cite{luo2018discriminability}, which designs a loss optimized to stir image captioning towards describing the differences between two images. \cite{vedantam2017context} suggest a transmitter-emitter (ES) architecture. \cite{andreas-klein-2016-reasoning,vedantam2017context,cohn2018pragmatically} generates pragmatically informative captions, by decoding general captioning models, at testing time, to produce captions that discriminate target images from a given set of distractor images. 

\paragraph{Zero-Shot Image Captioning.} RE methods can be viewed as a special case of image captioning methods. Recent work on open-world image captioning combines the abilities of two large pre-trained models: CLIP and GPT. \cite{su2022language} suggests regularizing sequences produced by GPT2 to be semantically related to a given image with a CLIP-induced score. Concurrently, \cite{tewel2022zero} proposed a similar technique that relies on gradient update and optimization over context cache. This improves accuracy but dramatically slows down inference. In later work, \cite{tewel2022zerocap} improve efficiency and inference speed by updating the context of a pseudo-token over different iterations in which the model generates full sentences. \cite{mokady2021clipcap} suggests producing meaningful captions by initializing GPT2 with visual prefix embeddings, which is learned by employing a simple mapping network from CLIP embedding space to GPT2 space. \cite{hessel2021clipscore} proposed to formalize the CLIP score as a new standard metric for image captioning.

\paragraph{Referring Expressions Generation (REG).} As far as we know, the current state-of-the-art approach in Referring Expression Generation (REG) is presented in \cite{schuz2021decoupling}. Their method involves incorporating pragmatic reasoning into context-agnostic generation models during inference. To generate pragmatically informative captions, they decode general captioning models at test time, producing captions that discriminate the target image from a set of distractor images. Decoding methods include criteria such as likelihood (Beam Search) and informativity (RSA decoding) \cite{cohn2018pragmatically}. Although some models achieve state-of-the-art (SotA) results on certain subsets of RefCOCO/+/g, there is no single model that consistently outperforms others across the board. A key difference is that our approach is designed to perform on objects within the same visual scene, rather than a curated set of distractors. 

\cite{tanaka2019generating} studies REs from the standpoint of object saliency. It has been observed that salient objects can be referred to using short and simple phrases, whereas less salient objects require more complex descriptions that often involve relationships with other objects within the scene. While our work does not specifically target this aspect, we draw comparisons with this baseline due to its zero-shot REG framework. Another work by \cite{liu2017referring, liu2020attribute} explicitly learns visual attributes and uses it as a supervision signal for REG-REC modules. %
Recent work \cite{huang2022unified} achieves impressive results, but their approach is supervised and works in domain. Our focus is on out-of-domain generalization. 

\section{Workflows for REs generation and comprehension }\label{sec:workflow}
Visual referring expression involves two complementary tasks. First, \textit{generation}, where a speaker module is given an image and a bounding box of a target object and has to create a natural language expression that refers to that object. Second, \textit{comprehension} where a listener module parses the RE, with the goal of selecting the correct object in a given image.

There are two main strategies for training these modules. The first strategy is to pre-train a listener, freeze it, then use it as a frozen evaluator to measure the quality of predicted REs (e.g., \cite{luo2017comprehension,schuz2021decoupling}). Here, since the listener is fixed, it is used to calculate a static score, like box-selection accuracy, which can be readily used for training the speaker. The second approach is to train both modules jointly \cite{huang2022unified,vered2019joint}. This raises two main difficulties. First, since the generated language is discrete, passing gradients from the listener to the speaker is non-trivial and involves approximated optimization like using a Gumbel softmax or straight-through \cite{vered2019joint}.  Second, unless restricted, the two modules tend to drift away from natural language and pass information that is unintelligible to people \cite{vered2019joint}.
To alleviate this issue, some researchers use language quality metrics like BLEU against a ground truth set. Unfortunately, these measures tend to be highly insufficient \cite{callison2006re,anderson2016spice}. In all these cases, methods are trained on paired data of images, boxes, and ground-truth referring expressions collected from human raters. 
A potential issue with these workflows is that they tend to be limited to the distribution of the data they are trained on. Indeed, our experiments below show that when tested on new data, they may collapse and yield very low accuracy. 

How can we progress towards open-world referring expression generation?
    We wish to provide dataset-agnostic models that can provide referring expressions that are both natural and informative even for images outside the training distribution. To this end, we propose to use large pre-trained image captioning models \cite{radford2021learning}. These models are trained on massive web datasets and, as such, cover the long tail of visual and semantic content.

\begin{figure}[t]
    \centering
    \includegraphics[width=0.99\linewidth, trim={0 0 4in 0}, clip]{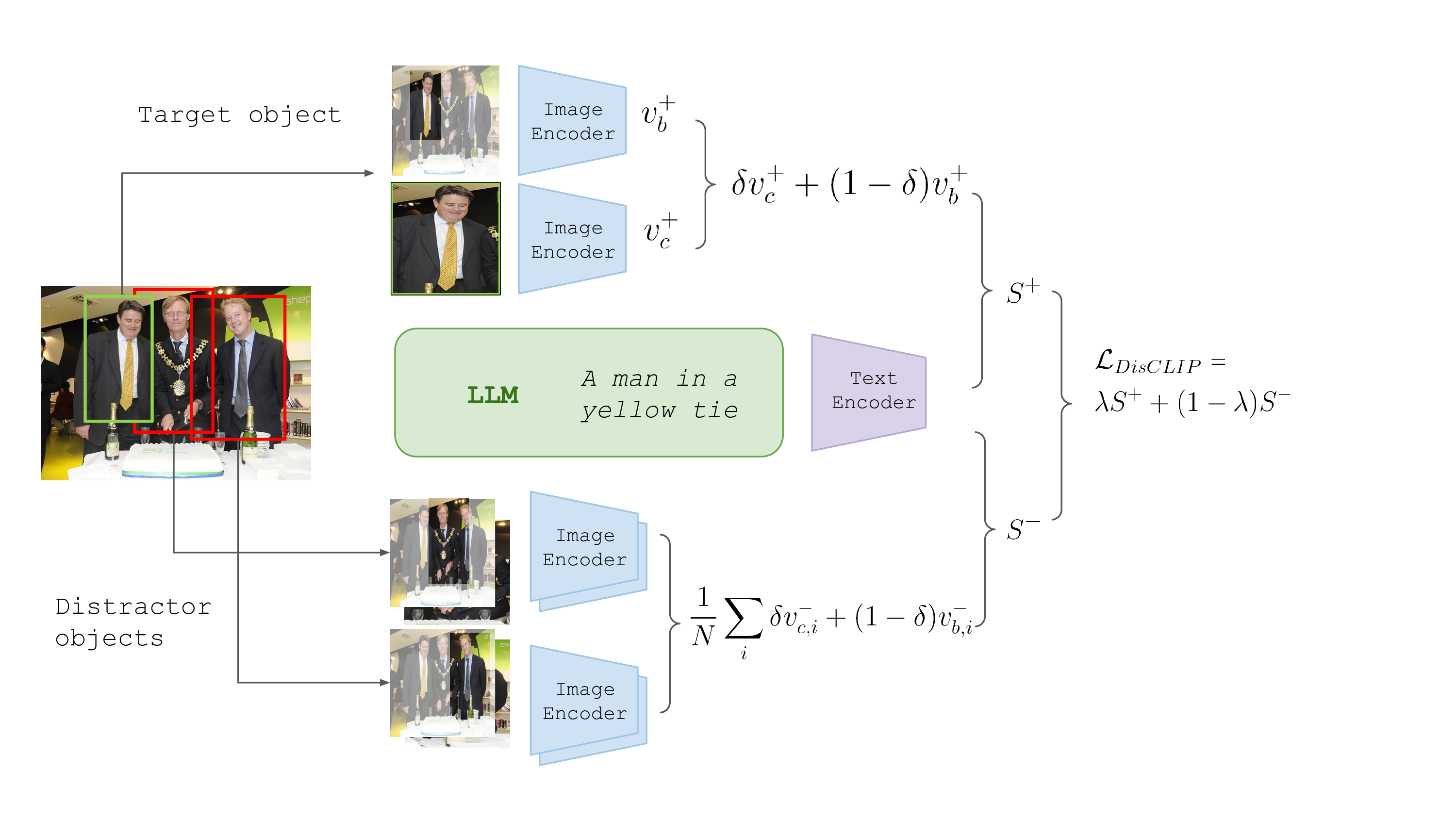}
    \caption{\small {\textbf{DisCLIP architecture for REG.} DisCLIP score encourages the language model (green) to generate text that is semantically related to the target object. The output sequence of the LM is encoded by CLIP's text encoder (purple). CLIP image encoder (blue) is used to encode representations of the target object ($v^+_{c}, v^+_{b}$) as well as the set of other objects in the scene ($v^-$). At each timestep, we maximize CLIP similarity with the target object and minimize similarity with a set of distractors.}} \vspace{-0.1in}
    \label{fig:method}
\end{figure}

\section{Model}

DisCLIP model is composed of two branches (Fig 2): a language branch where a \ac{LLM} generates a sequence of words (Fig. \ref{fig:method} green box), and a visual branch that guides generation to be close to an input image in a visual-semantic space. In an iterative process, we maximize the similarity \cite{hessel2021clipscore} between the generated sequence at a timestep $x_{<t}$ to \textit{the target region} in the image, and minimize the similarity to a set of distractors regions (namely, other objects). Our work is closely related to \cite{su2022language}, who put forward a similar process for zero-shot image captioning.

Let $x_{<t}$ be a sequence generated by an LLM at time $t$. Given input image $\mathcal{I}$, and $V^{(k)}$ (top $k$) candidate tokens from the LM, the probability of candidate token $v$ is computed as
\begin{equation}
    f(v|\mathcal{I}, x_{<t}, V^{(k)}) = \frac{e^{CLIP(I, [x_{<t}:v])}}{\sum_{z\in V^(k)} e^{CLIP(I, [x_{<t}:z])}} \quad ,\\ %
    \label{eq:magic}
\end{equation}
[:] denotes the concatenation operation, s.t. $x_{<t}:v$ represents the generated sequence so far, together with the current token. $CLIP(I,[x_{<t}:v])$ is the CLIP similarity score of an image $I$ and text $[x_{<t}:v]$. In our case, given an image $\mathcal{I}$ containing $n$ objects $O = \{ o_1, \ldots, o_n \}$ , we require that the generated sequence maximize the CLIP similarity with a target object $O^+$, while minimizing CLIP similarity to a set of distractors $O^- = \{ o^-_1, \ldots, o^-_{n-1} \}$. The total score is defined as %

\begin{equation}
    \mathcal{L}_{DisCLIP} =\lambda \bigg(
\overbrace{CLIP(O^{+}, [x_{<t}:v])}^{S^+} 
\bigg) + (1 - \lambda) \bigg( \frac{_{-1}}{^N} \sum_{i}
\overbrace{CLIP(O^{-}_{i}, [x_{<t}:v])}^{S_i^-}  \bigg) .
    \label{eq:disclip}
\end{equation} 
The hyper-parameter $\lambda \in [0,1]$ controls how strongly the negative set affects generation. When  $\lambda = 1$, negatives have no effect at all, and smaller values are expected to create increasingly discriminative text. 
The full objective includes terms designed for maintaining language fluency and consistency with the context tokens. For clarity, we refer to these terms as $\mathcal{L}_{lang}$, and describe them in detail in the Appendix \ref{sec:supp_loss}. 
\begin{equation}
    v = \argmax \Big \{ \mathcal{L}_{lang} + \beta \cdot \mathcal{L}_{disCLIP} \; \Big \}.
\end{equation}
Finally, the hyperparameter $\beta$ controls the trade-off between language weighs and the \disclip{} vision score.

\paragraph{Representing boxes.} In contrast with standard captioning, RE text-generating task has to: (i) describe a specific object in the scene instead of the entire image. This is challenging since CLIP was trained on image-level descriptions. (ii) The generated text should be contextual, which requires gathering information about the rest of the objects in the scene. This impacts how we experiment with the visual representation of the objects. 

To capture both local and global information we create different representations for each object in the scene. The first is simply a crop of the object's box, and the second is a blurred version of the image, except for the target region. We discuss other representations in Appendix \ref{sec:ablations}. 

Object representations are passed to the CLIP image encoder and used to compute the similarity to the generated text at time $t$. For the set of negatives, we sum over the similarity scores $S_i$, 
\begin{equation}
    S_i = \delta \cdot Blur(O_i) + (1 - \delta) \cdot Crop(O_i)\quad,
\end{equation}
where $\delta$ controls the trade-off between the two representations, as illustrated in Fig \ref{fig:method}.

\section{Experiments} \label{sec:exp}
We evaluate \disclip{} and the baselines in several experimental setups. 
We begin by showing out-of-domain performance measured using a pre-trained listener on three datasets: Flickr30k-Entities, RefCLEF, and RefGTA. Next, we put forward human evaluation results on our generated REs compared to the baselines. We also show that \disclip{} performs reasonably well in in-domain benchmarks compared to supervised methods. To encourage future research and reproducibility, we will make our source code publicly available.

\paragraph{Data.}

We used the following datasets. Since our method does not require training, we only used the validation and text splits in evaluations.
\noindent\textbf{(1) RefCOCO }\cite{kazemzadeh2014referitgame} contains 142,209 referring expressions for 50,000 objects in 19,994 images. 
\textbf{(2) RefCOCO+} \cite{kazemzadeh2014referitgame} contains 141,564 referring expressions for 49,856 objects in 19,992 images. This dataset focuses on objects' appearance, rather than spatial relations. In both RefCOCO and RefCOCO+, Test A contains references to humans, and Test B references to other object types. 
\textbf{(3) RefCOCOg (Google RefExp)} \cite{mao2016generation} contains 85,474 referring expressions for 54,822 objects in 26,711 images and contains longer and more complex expressions. 
\textbf{(4) RefCLEF} (ReferIt) \cite{kazemzadeh-etal-2014-referitgame} 
A dataset containing \textit{complex} photographs of real-world cluttered scenes. 10K test images, with \~ 60K references in the train/val set and 60,105 in the test set.  
\textbf{(5) RefGTA} \cite{tanaka2019generating}, contain 
synthetic images from the Grand Theft Auto (GTA) videogame. 6504 test images. All REs correspond to people, focusing on relations expressions.
\textbf{(6) Flickr30k-Entities} \cite{plummer2015flickr30k}, provides a comprehensive ground-truth correspondence between regions in images and phrases in captions. It contains 244K coreference chains, with 275K corresponding bounding boxes. We excluded ``group'' references (e.g. \textit{People are outside waving flags}), resulting in 1966 images and 4597 references in the validation set and 4601 in the test.  

\paragraph{Baselines.}

We compared our approach with the following baselines with their model that trained on RefCOCO+: 
\textbf{(1) Schutz et al. 2021} \cite{schuz2021decoupling} adopts an Emitter-Suppressor (ES) framework of \cite{vedantam2017context}. A speaker (E) models a caption for a target image $I_t$ in conjunction with a listener function (S) that rates how discriminative is the utterance with regard to a distractor image, $\lambda$ is a parameter that weighs the suppressor. We compare with  $\lambda = 0.5$ for his best model.
\textbf{(2) Tanaka et al. 2019 \cite{tanaka2019generating}} suggested an end-to-end training for  encoder decoder. Based on low-level visual representations as the input, various aspects of the task are modeled jointly, e.g. lexicalization and content selection.
\textbf{(3) Licheng Yu et al. 2017 \cite{yu2017joint}} proposed an end-to-end trained listener-speaker for RE task. He also added a discriminative reward-based module (reinforcer) to guide the sampling of more discriminative expressions and further improve his final model.\vspace{-0.15in}

\paragraph{Evaluation metrics.}
Standard evaluation metrics for REs like BLUE or CIDER \cite{schuz2021decoupling, su2022language} focus on agreement with ground-truth expressions. In the case of open-text generation, these metrics do not reflect true performance because LLMs produce rich natural sentences whereas GT phrases tend to be terse. To address this, we use two evaluation approaches: human raters and a frozen REC model -- a ``listener''. We follow the protocol in \cite{subramanian2022reclip, kamath2021mdetr} and measure listener accuracy as the percentage of instances for which the predicted box whose IoU with the ground-truth box is at least 0.5, a standard metric used to evaluate RE methods. For consistency with previous works in the field, we also report standard language metrics, provided in Appendix~\ref{sec:supp_language}.

\begin{table}[t]
    \centering
    \resizebox{0.99\textwidth}{!}{%
    \begin{tabular}{lp{2cm}cccccccc}
    && \multicolumn{2}{c}{RefClef} && \multicolumn{2}{c}{RefGTA }  && \multicolumn{2}{c}{Flickr30 Entities }   \\
    \cmidrule(lr){3-4} \cmidrule(lr){6-7} \cmidrule(lr){9-10} 
    & trained on & Test A & Test B & & Val & Test && Val & Test\\
    \midrule 
    \textbf{\textsc{Supervised methods}} &  &  & & & &  \\
    Schutz et al.\cite{schuz2021decoupling} & refCOCO+  & 26.0 & 18.2 & & 11.5 & 11.8 && 31.7 & 32.0  \\
    Tanaka et al. \cite{tanaka2019generating} & refCOCO+ & 27.0 & 33.4 & &  52.5 & 53.2 && 34.6 & 39.6 \\
     Licheng Yu et al. \cite{yu2017joint}  & refCOCO+ & 38.0 & 41.4 && 31.2 & 31.8 && 50.9 & 49.0\\
    \midrule
    \textbf{\textsc{Open-Vocabulary}} &  &  & & & &  \\
    \disclip{} (ours) + ReCLIP  \cite{subramanian2022reclip} & & 66.2 & 68.6  & & 58.0 & 56.9 && 77.9 & 78.8\\
    \disclip{}-HPT (ours) + ReCLIP  \cite{subramanian2022reclip} & & \textbf{83.4} & \textbf{ 85.4}    & & \textbf{73.4} & \textbf{73.6} & & \textbf{89.2} & \textbf{91.2}\\
    \bottomrule \\
    \end{tabular}
    }
    \caption{ \normalsize {\textbf{Out-of-domain generalization.}
    Listener accuracy on three different datasets,
    RefClef, RefGTA, and Flickr30k-Entities. }} 
    \label{tab:ood}
\end{table}

\section{Results}
\paragraph{Out-of-domain generalization} We now evaluate all methods in an out-of-domain setup. We trained the baseline methods on RefCOCO+, which capture attribute-based referrals. \disclip{} requires no training, but we tuned its hyperparameters $\delta$ and $\lambda$ on a subset  of 200 random samples from the validation split, see Figure \ref{fig:tuning}. In the evaluations below, we used the "natural" listener that is "paired" with the speaker in the sense that the listener was used either when training or evaluating the speaker in their original papers. 

\begin{table}[b]
    \centering
    \scriptsize
    \resizebox{0.99\textwidth}{!}{%
    \begin{tabular}{lp{2cm}cccccccc}
    && \multicolumn{2}{c}{RefClef} && \multicolumn{2}{c}{RefGTA }  && \multicolumn{2}{c}{Flickr30 Entities }   \\ 
    \cmidrule(lr){3-4} \cmidrule(lr){6-7} \cmidrule(lr){9-10} 
    & trained on & Test A & Test B & & Val & Test && Val & Test\\ 
    \midrule 
    GT RefExp &  & 65.5 & 64.4 && 40.3 & 40.6 && 72.6 & 73.9  \\ \hline
    Schutz et al.\cite{schuz2021decoupling} & refCOCO+  & 34.8 & 26.4 & & \textbf{40.8} & \textbf{40.9} && \textbf{40.7} & \textbf{40.6}  \\
    Tanaka et al. \cite{tanaka2019generating} & refCOCO+ & 22.8 & 20.4 & &  38.9 & 40.2 && 32.0 & 31.1 \\
     Licheng Yu et al. \cite{yu2017joint}  & refCOCO+ & 27.6 & 22.0 && 24.8 & 25.2 && 31.8 & 31.1\\
    \midrule
    \disclip{} (ours) && 35.0 &29.8  && 33.0 & 32.6 && 37.0 & 36.7\\
    \disclip{}-HPT (ours) && \textbf{36.2} & \textbf{30.8} && 33.9 & 33.3 & & 36.3 &  35.9\\
    \bottomrule \\
    \end{tabular}
    }
\caption{Evaluation of OOD with an independent listener module (mDETR)}\label{tab:mdetr}
\end{table}
\noindent Table \ref{tab:ood} shows results on the out-of-domain datasets RefCLEF, RefGTA, and Flickr30k-Entities, where \disclip{} significantly outperforms the baselines methods.

\paragraph{Independent pre-trained listener.}
The performance degradation of baselines observed in Table \ref{tab:ood} might result from the domain shift that the  listener (the REC model) experiences, rather than the speaker -- which is our prime interest. We further test a single pre-trained REC model as a common listener to evaluate all  different ``speakers'', in an identical way. For that listener, we choose mDETR  \cite{kamath2021mdetr}. It is an end-to-end modulated detector that detects objects in an image conditioned on a raw text query.
The results are presented in Table \ref{tab:mdetr}. DisCLIP outperforms the baselines on the RefClef dataset and is  competitive on RefGTA and Flickr30k entities. 

\begin{figure}[t]
    \centering
    \includegraphics[width=\linewidth, trim={0 15in 0 0}, clip]{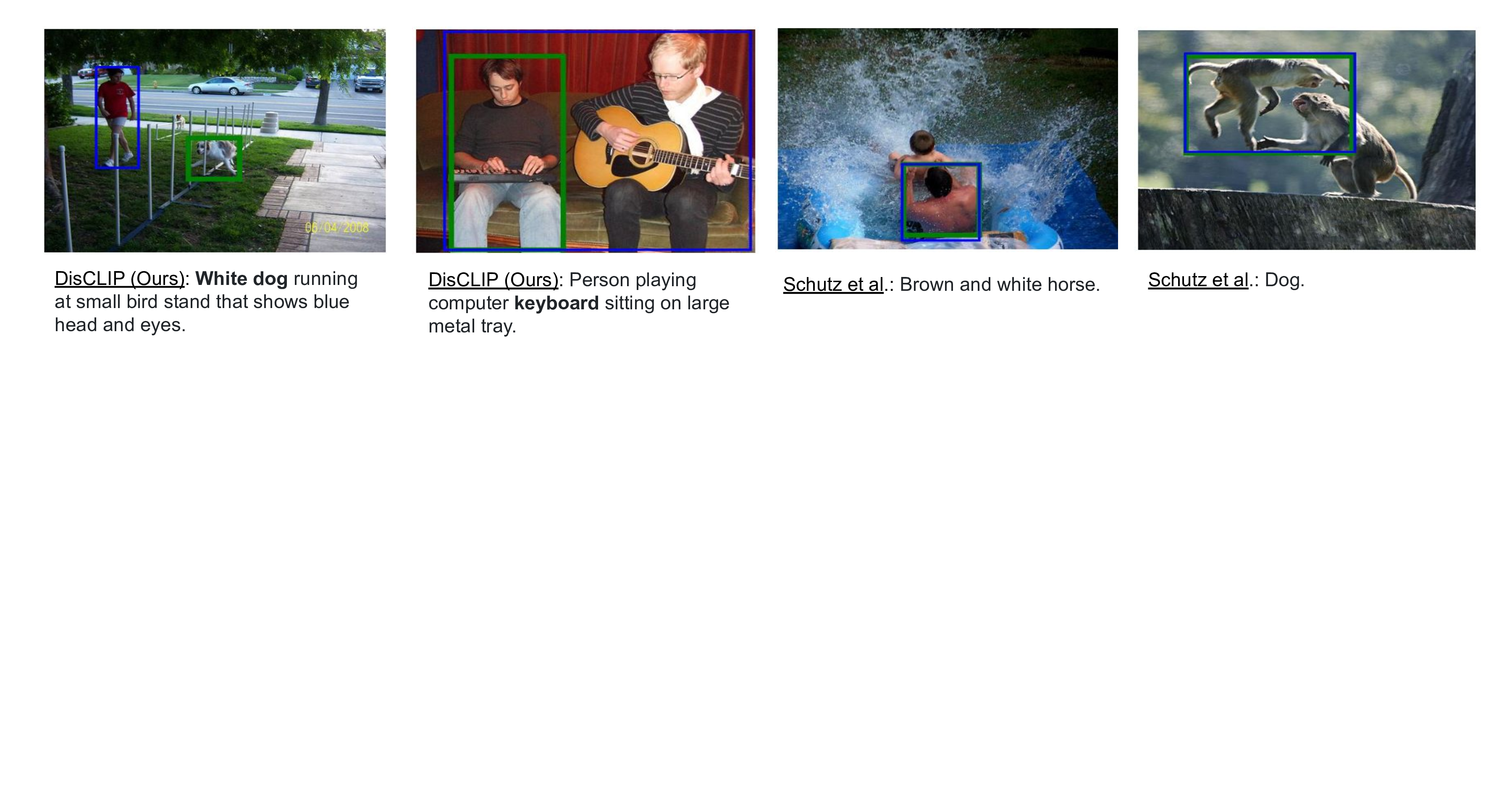}
    \caption{\small{An mDETR listener fails more with long natural sentences. In many cases, mDETR predicts a box (blue) that has a high overlap with GT box (green), even when the captions are completely unrelated to the image like in the two examples on the right. On the other hand, it often misses valid clues in the textual descriptions. }}
    \label{fig:mdetr}
\end{figure}

To understand this difference, we note that mDETR was fine-tuned on RefCOCO/+/g. Presumably, it became tuned to short sentences and perform worse on rich natural sentences.  
Indeed, from a qualitative error analysis, we find that mDETR makes more mistakes with long sentences, potentially causing a bias against disCLIP and favoring the baselines. See qualitative examples in Fig. \ref{fig:mdetr}.

\paragraph{Human evaluations on OOD REs.} 
Given the above limitations of out-of-the-box listeners as evaluators, as well as traditional language metrics, we follow up with evaluation by human raters. 
We generated REs for 100 random samples from three out-of-domain datasets, and sent each RE to three unique raters. Given a textual description (generated by us or the baselines), participants are asked to choose one out of $n$ candidate boxes that best matches the RE (details in Appendix \ref{sec:supp_mturk}). Table \ref{tab:human_eval} shows that human raters prefer phrases generated by \disclip{} model, across all out-of-domain datasets by a large margin. Our method generates more diverse and natural phrases compared to baseline methods as shown by the qualitative examples in Sec. \ref{sec:sup_mturk_qual} of the Appendix. 
 
\begin{table}[hb]
    \centering
    \scriptsize
    \resizebox{0.99\textwidth}{!}{%
    \begin{tabular}{lp{2cm}ccccc}
    && \multicolumn{1}{c}{RefClef} && \multicolumn{1}{c}{RefGTA }  && \multicolumn{1}{c}{Flickr30 Entities }   \\
    \cmidrule(lr){3-3} \cmidrule(lr){5-5} \cmidrule(lr){7-7} 
    & trained on & Test A && Val && Test\\
    \midrule 
    Schutz et al.\cite{schuz2021decoupling} & refCOCO+  & 31.0 && 36.0  && 43.1    \\
    Tanaka et al. \cite{tanaka2019generating} & refCOCO+ & 6.0 && 46.0 && 19.0  \\
     Licheng Yu et al. \cite{yu2017joint}  & refCOCO+ & 14.0  && 26.0 && 29.0 \\
    \midrule
    \disclip{} (ours)  && \textbf{46.3} && \textbf{49.0} && \textbf{45.7}  \\
    \bottomrule \\
    \end{tabular}
    }
    \caption{\small{Human evaluation on the out-of-domain datasets. On average, DisCLIP outperform the baseline methods by a margin of 29.3\% on RefClef, 13\% on RefGTA and 15.3\% on Flickr30-Entities.}}\vspace{-0.15in}
    \label{tab:human_eval}
\end{table}

\begin{table}[hb]
    \centering
    \resizebox{0.99\textwidth}{!}{%
    \begin{tabular}{lccccccccccc}
    & & \multicolumn{3}{c}{In domain} &   \multicolumn{5}{c}{GT Label shift} \\
    \cmidrule(lr){3-5} \cmidrule(lr){6-10} 
    & & \multicolumn{3}{c}{RefCOCO+} & \multicolumn{3}{c}{RefCOCO} &  \multicolumn{2}{c}{RefCOCOg} \\
    \cmidrule(lr){3-5} \cmidrule(lr){6-8} \cmidrule(lr){9-10} 
    & trained on & Val  & Test A & Test B &  Val & Test A  & Test B & Val & Test\\
    \midrule 
    \textbf{\textsc{Supervised methods}} &  &  & & & & & & \\
    Schutz et al. \cite{schuz2021decoupling} & refCOCO+ & 58.3 & 68.4 & 48.2 &  58.2 & 68.4 & 48.1 & 62.1 & 62.5 \\
    
    Tanaka et al. \cite{tanaka2019generating} & refCOCO+  & 65.8 & 70.9 & 62.5 & 65.8 & 70.9 & 62.2 & 72.0 & 71.4\\
    
    Licheng Yu et al. \cite{yu2017joint} & refCOCO+  & \textbf{79.2} & \textbf{82.9} & 75.0 & \textbf{79.1} &  \textbf{82.9} & \textbf{74.6} &  \textbf{86.1} & 85.7  \\ 
    \midrule
    \textbf{\textsc{Open-Vocabulary}} &  &  & & & &  & & \\
    \disclip{} (ours) + ReCLIP \cite{subramanian2022reclip} & & 67.3 &  70.1 &  64.7 &  67.2 & 70.1  & 64.5 & 72.8 & 74.9 \\ 
    \disclip{}-HPT (ours) + ReCLIP \cite{subramanian2022reclip} &  & 78.6 & 80.2 & \textbf{77.2} & 76.5 & 80.2 & 73.7 & 85.1 & \textbf{86.5} \\
    \bottomrule \\
    \end{tabular}
    }
    \caption{\small{\textbf{In-domain accuracy} of models tested on three variants of RefCOCO. Each method uses paired (jointly trained) speaker and listener. All datasets have the same distribution of images, but GT labels shift between datasets.}} \label{tab:in-dist} \vspace{-0.15in}
\end{table}

\begin{figure}[hbpt]
    \centering
    \includegraphics[width=\linewidth, trim={0 2in 0 0}, clip]{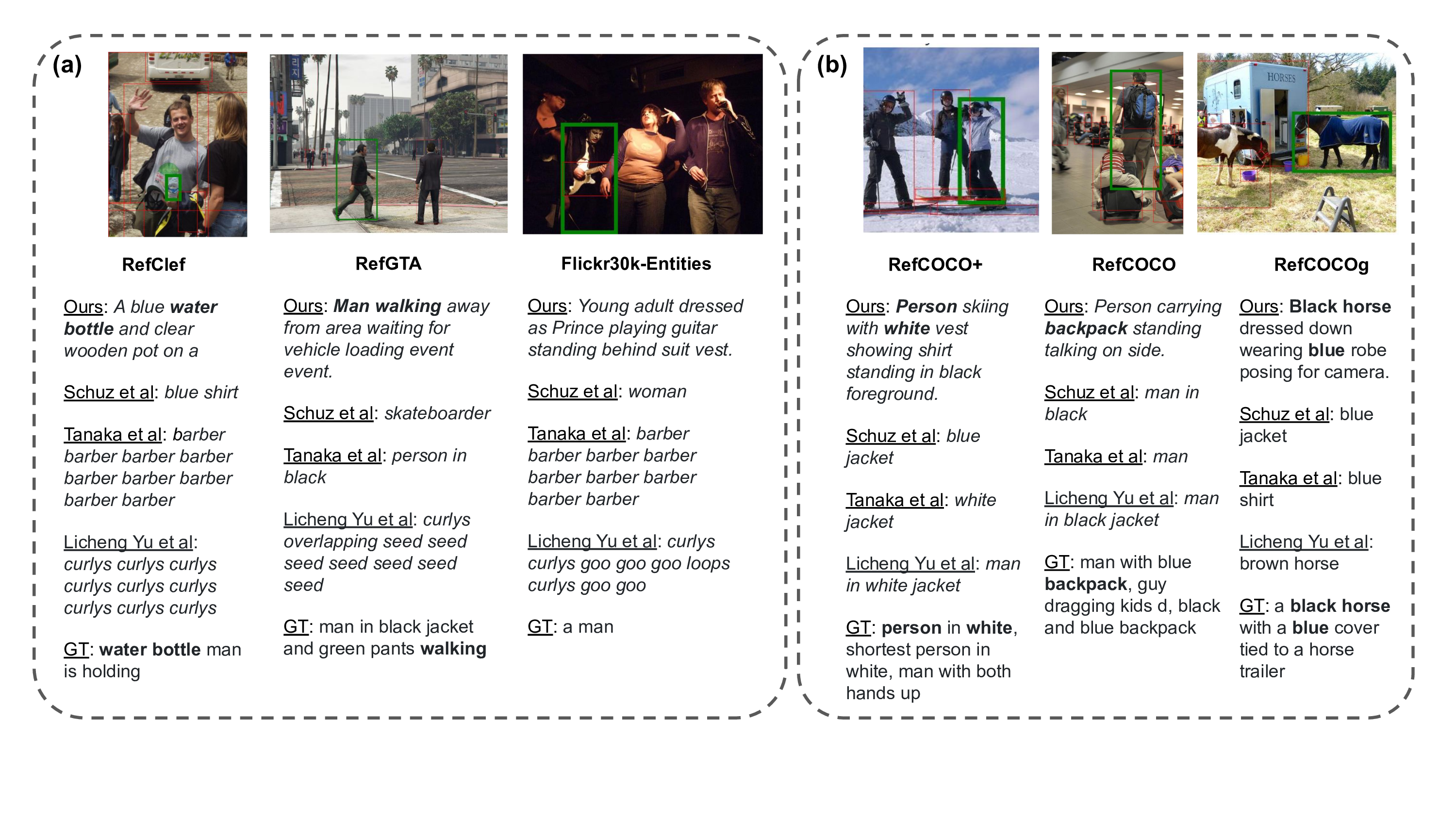} 
    \caption{\small{\textbf{Qualitative results} on out-of-domain datasets (panel a), and in-domain datasets (panel b)}} \label{fig:qual}
\end{figure}

\paragraph{In-domain referring expression.}
\disclip{} is designed for out-domain and open-vocabulary setup. For completeness, we also tested its accuracy on in-domain datasets RefCOCO/+/g in Table \ref{tab:in-dist}. Baseline models had both their listener and speaker trained on RefCOCO+ (attribute-based REs). There are also versions of the baseline models that were trained on RefCOCO, but since it is focused on spatial phrases, this comparison is less relevant to our task, which focuses on attribute-based REs. Table \ref{tab:in-dist} shows that \disclip{} stays competitive with the supervised baselines on all the in-domain datasets. Qualitative examples from both in and out-of-domain are shown in Fig. \ref{fig:qual}. \vspace{-0.15in}

\paragraph{Limitations.}
DisCLIP is successful, but it is also important to address its limitations. First, CLIP has notorious poor sensitivity to spatial relations. As a result, the expressions generated by our model use attribute-based REs rather than relation-based REs, like ``bike on right''.
Second, Our language generation is very simple, generating the expression token by token. It is appealing that smarter models for expression generation may improve performance. Given that \disclip{} does not rely on any training or fine-tuning procedures, using better  foundation models in the future  is expected to yield better REG using similar \disclip{} inference.

\section{Conclusion}
In this work, we present a novel method, named \disclip{}, to generate discriminative referring expressions in an open-world setting. Instead of training a model for one specific dataset, we leverage large pre-trained foundation models (CLIP, GPT2).
\disclip{} achieves significant improvement over baseline models trained on different datasets, showing robustness to the domain shift occurring across datasets.

\section*{Acknowledgements} This study was funded by a grant to GC from the Israel
Science Foundation (ISF 737/2018), and by an equipment grant to GC and Bar-Ilan University from the Israel Science Foundation (ISF 2332/18). Lior Bracha is supported by a PhD fellowship in data science from the Israeli national council of higher education.

\bibliography{ms}

\begin{thebibliography}{10}

\bibitem{andreas2016reasoning}
Jacob Andreas and Dan Klein.
\newblock Reasoning about pragmatics with neural listeners and speakers.
\newblock {\em arXiv preprint arXiv:1604.00562}, 2016.

\bibitem{kazemzadeh2014referitgame}
Sahar Kazemzadeh, Vicente Ordonez, Mark Matten, and Tamara Berg.
\newblock Referitgame: Referring to objects in photographs of natural scenes.
\newblock In {\em Proceedings of the 2014 conference on empirical methods in
  natural language processing (EMNLP)}, pages 787--798, 2014.

\bibitem{andreas-klein-2016-reasoning}
Jacob Andreas and Dan Klein.
\newblock Reasoning about pragmatics with neural listeners and speakers.
\newblock In {\em Proceedings of the 2016 Conference on Empirical Methods in
  Natural Language Processing}, pages 1173--1182, Austin, Texas, nov 2016.
  Association for Computational Linguistics.

\bibitem{vedantam2017context}
Ramakrishna Vedantam, Samy Bengio, Kevin Murphy, Devi Parikh, and Gal Chechik.
\newblock Context-aware captions from context-agnostic supervision.
\newblock In {\em Proceedings of the IEEE Conference on Computer Vision and
  Pattern Recognition}, pages 251--260, 2017.

\bibitem{vered2019joint}
Gilad Vered, Gal Oren, Yuval Atzmon, and Gal Chechik.
\newblock Joint optimization for cooperative image captioning.
\newblock In {\em Proceedings of the IEEE/CVF International Conference on
  Computer Vision}, pages 8898--8907, 2019.

\bibitem{luo2018discriminability}
Ruotian Luo, Brian Price, Scott Cohen, and Gregory Shakhnarovich.
\newblock Discriminability objective for training descriptive captions.
\newblock In {\em Proceedings of the IEEE Conference on Computer Vision and
  Pattern Recognition}, pages 6964--6974, 2018.

\bibitem{radford2021learning}
Alec Radford, Jong~Wook Kim, Chris Hallacy, Aditya Ramesh, Gabriel Goh,
  Sandhini Agarwal, Girish Sastry, Amanda Askell, Pamela Mishkin, Jack Clark,
  et~al.
\newblock Learning transferable visual models from natural language
  supervision.
\newblock In {\em International Conference on Machine Learning}, pages
  8748--8763. PMLR, 2021.

\bibitem{frank2012predicting}
Michael~C Frank and Noah~D Goodman.
\newblock Predicting pragmatic reasoning in language games.
\newblock {\em Science}, 336(6084):998--998, 2012.

\bibitem{luo2017comprehension}
Ruotian Luo and Gregory Shakhnarovich.
\newblock Comprehension-guided referring expressions.
\newblock In {\em Proceedings of the IEEE Conference on Computer Vision and
  Pattern Recognition}, pages 7102--7111, 2017.

\bibitem{liu2020attribute}
Jingyu Liu, Wei Wang, Liang Wang, and Ming-Hsuan Yang.
\newblock Attribute-guided attention for referring expression generation and
  comprehension.
\newblock {\em IEEE Transactions on Image Processing}, 29:5244--5258, 2020.

\bibitem{mao2016generation}
Junhua Mao, Jonathan Huang, Alexander Toshev, Oana Camburu, Alan Yuille, and
  Kevin Murphy.
\newblock Generation and comprehension of unambiguous object descriptions.
\newblock In {\em CVPR}, 2016.

\bibitem{hu2016natural}
Ronghang Hu, Huazhe Xu, Marcus Rohrbach, Jiashi Feng, Kate Saenko, and Trevor
  Darrell.
\newblock Natural language object retrieval.
\newblock In {\em Proceedings of the IEEE conference on computer vision and
  pattern recognition}, pages 4555--4564, 2016.

\bibitem{yu2017joint}
Licheng Yu, Hao Tan, Mohit Bansal, and Tamara~L Berg.
\newblock A joint speaker-listener-reinforcer model for referring expressions.
\newblock In {\em Proceedings of the IEEE Conference on Computer Vision and
  Pattern Recognition}, pages 7282--7290, 2017.

\bibitem{cohn2018pragmatically}
Reuben Cohn-Gordon, Noah Goodman, and Christopher Potts.
\newblock Pragmatically informative image captioning with character-level
  inference.
\newblock {\em arXiv preprint arXiv:1804.05417}, 2018.

\bibitem{su2022language}
Yixuan Su, Tian Lan, Yahui Liu, Fangyu Liu, Dani Yogatama, Yan Wang, Lingpeng
  Kong, and Nigel Collier.
\newblock Language models can see: Plugging visual controls in text generation.
\newblock {\em arXiv preprint arXiv:2205.02655}, 2022.

\bibitem{tewel2022zero}
Yoad Tewel, Yoav Shalev, Roy Nadler, Idan Schwartz, and Lior Wolf.
\newblock Zero-shot video captioning with evolving pseudo-tokens.
\newblock {\em arXiv preprint arXiv:2207.11100}, 2022.

\bibitem{tewel2022zerocap}
Yoad Tewel, Yoav Shalev, Idan Schwartz, and Lior Wolf.
\newblock Zerocap: Zero-shot image-to-text generation for visual-semantic
  arithmetic.
\newblock In {\em Proceedings of the IEEE/CVF Conference on Computer Vision and
  Pattern Recognition}, pages 17918--17928, 2022.

\bibitem{mokady2021clipcap}
Ron Mokady, Amir Hertz, and Amit~H Bermano.
\newblock Clipcap: Clip prefix for image captioning.
\newblock {\em arXiv preprint arXiv:2111.09734}, 2021.

\bibitem{hessel2021clipscore}
Jack Hessel, Ari Holtzman, Maxwell Forbes, Ronan~Le Bras, and Yejin Choi.
\newblock Clipscore: A reference-free evaluation metric for image captioning.
\newblock {\em arXiv preprint arXiv:2104.08718}, 2021.

\bibitem{schuz2021decoupling}
Simeon Sch{\"u}z and Sina Zarrie{\ss}.
\newblock Decoupling pragmatics: Discriminative decoding for referring
  expression generation.
\newblock In {\em Proceedings of the Reasoning and Interaction Conference
  (ReInAct 2021)}, pages 47--52, 2021.

\bibitem{tanaka2019generating}
Mikihiro Tanaka, Takayuki Itamochi, Kenichi Narioka, Ikuro Sato, Yoshitaka
  Ushiku, and Tatsuya Harada.
\newblock Generating easy-to-understand referring expressions for target
  identifications.
\newblock In {\em Proceedings of the IEEE/CVF International Conference on
  Computer Vision}, pages 5794--5803, 2019.

\bibitem{liu2017referring}
Jingyu Liu, Liang Wang, and Ming-Hsuan Yang.
\newblock Referring expression generation and comprehension via attributes.
\newblock In {\em Proceedings of the IEEE International Conference on Computer
  Vision}, pages 4856--4864, 2017.

\bibitem{huang2022unified}
Shijia Huang, Feng Li, Hao Zhang, Shilong Liu, Lei Zhang, and Liwei Wang.
\newblock A unified mutual supervision framework for referring expression
  segmentation and generation.
\newblock {\em arXiv preprint arXiv:2211.07919}, 2022.

\bibitem{callison2006re}
Chris Callison-Burch, Miles Osborne, and Philipp Koehn.
\newblock Re-evaluating the role of bleu in machine translation research.
\newblock In {\em 11th conference of the european chapter of the association
  for computational linguistics}, pages 249--256, 2006.

\bibitem{anderson2016spice}
Peter Anderson, Basura Fernando, Mark Johnson, and Stephen Gould.
\newblock Spice: Semantic propositional image caption evaluation.
\newblock In {\em Computer Vision--ECCV 2016: 14th European Conference,
  Amsterdam, The Netherlands, October 11-14, 2016, Proceedings, Part V 14},
  pages 382--398. Springer, 2016.

\bibitem{kazemzadeh-etal-2014-referitgame}
Sahar Kazemzadeh, Vicente Ordonez, Mark Matten, and Tamara Berg.
\newblock {R}efer{I}t{G}ame: Referring to objects in photographs of natural
  scenes.
\newblock In {\em Proceedings of the 2014 Conference on Empirical Methods in
  Natural Language Processing ({EMNLP})}, pages 787--798, Doha, Qatar, oct
  2014. Association for Computational Linguistics.

\bibitem{plummer2015flickr30k}
Bryan~A Plummer, Liwei Wang, Chris~M Cervantes, Juan~C Caicedo, Julia
  Hockenmaier, and Svetlana Lazebnik.
\newblock Flickr30k entities: Collecting region-to-phrase correspondences for
  richer image-to-sentence models.
\newblock In {\em Proceedings of the IEEE international conference on computer
  vision}, pages 2641--2649, 2015.

\bibitem{subramanian2022reclip}
Sanjay Subramanian, William Merrill, Trevor Darrell, Matt Gardner, Sameer
  Singh, and Anna Rohrbach.
\newblock Reclip: A strong zero-shot baseline for referring expression
  comprehension.
\newblock In {\em Proceedings of the 60th Annual Meeting of the Association for
  Computational Linguistics (Volume 1: Long Papers)}, pages 5198--5215, 2022.

\bibitem{kamath2021mdetr}
Aishwarya Kamath, Mannat Singh, Yann LeCun, Gabriel Synnaeve, Ishan Misra, and
  Nicolas Carion.
\newblock Mdetr-modulated detection for end-to-end multi-modal understanding.
\newblock In {\em Proceedings of the IEEE/CVF International Conference on
  Computer Vision}, pages 1780--1790, 2021.

\bibitem{papineni2002bleu}
Kishore Papineni, Salim Roukos, Todd Ward, and Wei-Jing Zhu.
\newblock Bleu: a method for automatic evaluation of machine translation.
\newblock In {\em Proceedings of the 40th annual meeting of the Association for
  Computational Linguistics}, pages 311--318, 2002.

\bibitem{vedantam2015cider}
Ramakrishna Vedantam, C~Lawrence~Zitnick, and Devi Parikh.
\newblock Cider: Consensus-based image description evaluation.
\newblock In {\em Proceedings of the IEEE conference on computer vision and
  pattern recognition}, pages 4566--4575, 2015.

\bibitem{lin2004automatic}
Chin-Yew Lin and Franz~Josef Och.
\newblock Automatic evaluation of machine translation quality using longest
  common subsequence and skip-bigram statistics.
\newblock In {\em Proceedings of the 42nd Annual Meeting of the Association for
  Computational Linguistics (ACL-04)}, pages 605--612, 2004.

\bibitem{elliott2013image}
Desmond Elliott and Frank Keller.
\newblock Image description using visual dependency representations.
\newblock In {\em Proceedings of the 2013 conference on empirical methods in
  natural language processing}, pages 1292--1302, 2013.

\bibitem{su2022contrastive}
Yixuan Su, Tian Lan, Yan Wang, Dani Yogatama, Lingpeng Kong, and Nigel Collier.
\newblock A contrastive framework for neural text generation.
\newblock {\em arXiv preprint arXiv:2202.06417}, 2022.

\end{thebibliography}
\bibliographystyle{unsrt}

\newpage 
\appendix
\Large{\textbf{Appendix}} \\
\normalsize

DisCLIP is a framework that generates referring expressions for objects in a scene by leveraging an LLM and CLIP. It guides a language model by iteratively aligning the generated text with the target region in the CLIP space. This appendix includes additional experimental results and analysis. To facilitate comparison with previous supervised methods, we report standard language metrics. Furthermore, we conduct an ablation study to examine the impact of object representation methods and hyper-parameters on robustness. Lastly, we provide qualitative examples where human raters either succeeded or failed in a Referential Expression Comprehension (REC) task when provided with descriptions generated by our model and the baseline methods.

\section{Language Metrics} \label{sec:supp_language}

Commonly used evaluation metrics for referring expressions (REs) include language metrics such as BLUE \cite{papineni2002bleu}, CIDEr \cite{vedantam2015cider}, ROUGE-L \cite{lin2004automatic}, and METEOR \cite{elliott2013image}. These metrics primarily assess the agreement between generated expressions and a set of ground-truth references. However, when it comes to open-text generation, these metrics may not be suitable since language models (LLMs) produce detailed natural sentences while ground-truth expressions tend to be terse. Nonetheless, for the sake of consistency with previous studies, we provide the results of standard language metrics in Table \ref{tab:lang}.

\begin{table}[th]
        \resizebox{\textwidth}{!}{%
        \begin{tabular}{lccccccccccc} \toprule
         & \multicolumn{11}{c}{\textsc{RefClef}} \\ \cmidrule(lr){2-12} 
         & \multicolumn{5}{c}{Test A (Human)} && \multicolumn{5}{c}{Test B (Objects)} \\
         \cmidrule(lr){2-7} \cmidrule(lr){8-12} 
        \multirow{-3}{*}{} & \multicolumn{1}{c}{$BLEU1 \uparrow$} & \multicolumn{1}{c}{$BLEU4 \uparrow$} & \multicolumn{1}{c}{$METEOR \uparrow$ } & \multicolumn{1}{c}{$CIDEr\uparrow$} & \multicolumn{1}{c}{$Rouge-L \uparrow$ } &  & \multicolumn{1}{c}{$BLEU1 \uparrow$} & \multicolumn{1}{c}{$BLEU4 \uparrow$} & \multicolumn{1}{c}{$METEOR \uparrow$ } & \multicolumn{1}{c}{$CIDEr\uparrow$} & \multicolumn{1}{c}{$Rouge-L \uparrow$} \\
        \cmidrule(lr){1-7} \cmidrule(lr){8-12} 
        Schutz et al.\cite{schuz2021decoupling} &  \textbf{0.226} & 0.000 & 0.080 & \textbf{0.330} & \textbf{0.215} && \textbf{0.183} & 0.000 & 0.066 & \textbf{0.200} & \textbf{0.151} \\
        Tanaka et al. \cite{tanaka2019generating} & 0.022 & 0.000 & 0.020 & 0.095 & 0.057 && 0.020  & 0.003 & 0.019 & 0.077 & 0.049 \\
        Yu et al. \cite{yu2017joint} & 0.046 & 0.000 & 0.035 & 0.109 & 0.084 && 0.031 & \textbf{0.004} & 0.028 & 0.085 & 0.062 \\ \midrule
        \textsc{DisCLIP (Ours)} & 0.123 & \textbf{0.005}  &\textbf{ 0.097 }& 0.088 & 0.158 && 0.120 & 0.000 & \textbf{0.090} & 0.063 & 0.144 \\
        \textsc{DisCLIP-HPT (Ours)} & 0.096 & 0.000 & 0.080 & 0.059 &  0.126 && 0.098 &  0.000 &  0.083 &  0.060 & 0.126 \\ 
        \bottomrule
        \end{tabular}}
        \resizebox{\textwidth}{!}{%
        \begin{tabular}{lccccccccccc} \toprule
         & \multicolumn{11}{c}{\textsc{RefGTA}} \\ \cmidrule(lr){2-12} 
         & \multicolumn{5}{c}{Validation} && \multicolumn{5}{c}{Test} \\
         \cmidrule(lr){2-7} \cmidrule(lr){8-12} 
        \multirow{-3}{*}{} & \multicolumn{1}{c}{$BLEU1 \uparrow$} & \multicolumn{1}{c}{$BLEU4 \uparrow$} & \multicolumn{1}{c}{$Meteor \uparrow$ } & \multicolumn{1}{c}{$CIDEr\uparrow$} & \multicolumn{1}{c}{$Rouge-L \uparrow$ } &  & \multicolumn{1}{c}{$BLEU1 \uparrow$} & \multicolumn{1}{c}{$BLEU4 \uparrow$} & \multicolumn{1}{c}{$Meteor \uparrow$ } & \multicolumn{1}{c}{$CIDEr\uparrow$} & \multicolumn{1}{c}{$Rouge-L \uparrow$} \\
        \cmidrule(lr){1-7} \cmidrule(lr){8-12} 
        Schutz et al.\cite{schuz2021decoupling}  & 0.078 & 0.018 & 0.072  & \textbf{0.161} & 0.192  && 0.076  & \textbf{0.019}  & 0.072  & \textbf{0.164} & 0.190   \\
        Tanaka et al. \cite{tanaka2019generating}  & 0.110 & 0.012  & 0.063  & 0.151 & 0.179   && 0.110 & 0.014 & 0.064  & 0.158 & 0.179  \\
        \textsc{Yu} et al. \cite{yu2017joint}  & 0.000 &   0.000 &  0.000 & 0.000 & 0.000 &&  0.000 &  0.000 & 0.000 & 0.000 & 0.000 \\ \midrule
        \textsc{DisCLIP (Ours)} & \textbf{0.307} & \textbf{0.018}  & \textbf{0.097}  & 0.070 & \textbf{0.234}    && \textbf{0.301}  & 0.017  & \textbf{0.096}  & 0.068  & \textbf{0.233}  \\
        \textsc{DisCLIP-HPT (Ours)}  & 0.196 & 0.004  & 0.079  & 0.048 & 0.162  && 0.196  & 0.004  & 0.078  & 0.047 & 0.161   \\ \bottomrule
        \end{tabular}} 
        \resizebox{\textwidth}{!}{%
        \begin{tabular}{lccccccccccc} \toprule
         & \multicolumn{11}{c}{\textsc{Flickr30k Entities}} \\ \cmidrule(lr){2-12} 
         & \multicolumn{5}{c}{Validation} && \multicolumn{5}{c}{Test} \\
         \cmidrule(lr){2-7} \cmidrule(lr){8-12} 
        \multirow{-3}{*}{} & \multicolumn{1}{c}{$BLEU1 \uparrow$} & \multicolumn{1}{c}{$BLEU4 \uparrow$} & \multicolumn{1}{c}{$Meteor \uparrow$ } & \multicolumn{1}{c}{$CIDEr\uparrow$} & \multicolumn{1}{c}{$Rouge-L \uparrow$ } &  & \multicolumn{1}{c}{$BLEU1 \uparrow$} & \multicolumn{1}{c}{$BLEU4 \uparrow$} & \multicolumn{1}{c}{$Meteor \uparrow$ } & \multicolumn{1}{c}{$CIDEr\uparrow$} & \multicolumn{1}{c}{$Rouge-L \uparrow$} \\
        \cmidrule(lr){1-7} \cmidrule(lr){8-12} 
        Schutz et al.\cite{schuz2021decoupling}     &  0.140 & 0.008 &  0.078 & \textbf{0.309} & 0.141  && \textbf{0.141} & 0.000 & 0.079 &  \textbf{0.329} &  0.190  \\
        Tanaka et al. \cite{tanaka2019generating}   &  0.024 & 0.001 & 0.027 & 0.148 & 0.066 && 0.022 & 0.000 & 0.024 & 0.150 & 0.061 \\
        \textsc{Yu} et al. \cite{yu2017joint}  &    0.028 &  0.000  &  0.030  & 0.141 &  0.068  && 0.024 & 0.000 & 0.027 & 0.139 & 0.063 \\ \midrule
        \textsc{DisCLIP (Ours)}  & 0.098  & 0.003 & \textbf{0.095} & 0.067 & \textbf{0.156} && 0.098  & \textbf{0.002} & \textbf{0.096}  & 0.068 & \textbf{0.157}    \\
        \textsc{DisCLIP-HPT (Ours)}   & 0.050    & 0.001   & 0.065  & 0.043 & 0.080 && 0.050 & 0.000 & 0.064  & 0.044 & 0.081 \\ \bottomrule
        \end{tabular}} \vspace{6pt}
\caption{Language Metrics for REG} \label{tab:lang}
\end{table}

\paragraph{Metrics for open-vocabulary text generation} 

In the context of unsupervised or open-text generation settings, several metrics have been proposed \cite{su2022contrastive}. These metrics are computed without relying on human annotation. One such metric is \textit{Relatedness to the image}, which assesses the distance between the image and the generated text using a retrieval-based approach. \textit{Language quality} is evaluated by estimating the \textit{perplexity} score of the generated caption, utilizing BERT. The perplexity score, which is the negative logarithm of the likelihood, reflects the level of uncertainty in the model's text predictions. Another metric, \textit{Diversity}, measures the vocabulary size and the percentage of novel sentences compared to the training set (\%Novel). 

\section{Ablation study}\label{sec:ablations}

\paragraph{Representing objects.} In our ablation study, we explored various methods for representing objects in the image. These methods include: (i) cropping the object's bounding box, (ii) blurring the entire image except the target region, and (iii) cropping with mirror padding for non-squared boxes, meant to maintain the proportions of the object.

Our findings indicate that a combination of cropping and blurring yields optimal results for this task. Our hypothesis is that this representation effectively encodes both local and global information. The act of cropping guides CLIP to focus on the target object, while the blurred surroundings provide valuable contextual cues. Moreover, since CLIP resizes the input image, cropped regions can sometimes become stretched, making it difficult to recover the object's semantics. In such cases, the blur representation proves useful as it maintains the objects' sizes and overall positioning of objects in the scene.

To balance between these two representations, we introduce the parameter $\delta$. Table~\ref{tab:ablation} compares several representation methods that we tested, using a fixed value of $\delta=0.5$ determined through hyper-parameter tuning.

\begin{table}[hbt!]
    \centering
    \resizebox{0.6\textwidth}{!}{
    \begin{tabular}{lccc} \toprule
    ReCLIP & RefClef - Test A & Flickr30k - Val \\ \midrule
    \disclip{}, crop-blur  &\ \textbf{67.4} & \textbf{77.9} \\
    \disclip{}, only blur  &  48.8 & 63.7  \\ 
    \disclip{}, only mirror  & 45.2 & 58.1 \\ 
    \disclip{}, only crop  &  52.2 & 61.7 \\ 
    \bottomrule \\
    \end{tabular} }
    \caption{Accuracy of ReCLIP listener given different representations of the target object. } 
    \label{tab:ablation}
\end{table}

\begin{table}[hbt!]
    \centering
    \resizebox{0.6\textwidth}{!}{
    \begin{tabular}{lccc} \toprule
    mDETR & RefClef - Test A & Flickr30k - Val \\ \midrule
    \disclip{}, crop-blur  &\ \textbf{35.0} & \textbf{37.0} \\
    \disclip{}, only blur  &  29.4 & 32.7  \\ 
    \disclip{}, only mirror  & 27.0 & 31.1 \\ 
    \disclip{}, only crop  &  30.4 & 34.2 \\ 
    \bottomrule \\
    \end{tabular}}
    \caption{Accuracy of mDETR listener given different representations of the target object.}
    \label{tab:ablation-mdetr}
\end{table}

\section{Robustness of the speaker}
In order to assess the robustness of the various speaker models, we aim to evaluate the extent of overfitting to the specific listener employed in the original paper. To achieve this, we decouple the speaker-listener pairs and measure accuracy across all possible combinations. Table \ref{tab:atificial-paring} presents the accuracy of each speaker (REC task) when paired with different listeners. The cells highlighted in blue indicate paired speaker-listener combinations, where the listener was either trained jointly with the respective speaker or used in the original paper. It is worth noting that all supervised listeners underwent training on refCOCO+.

\begin{table}[h!]
\centering
    \resizebox{0.9\textwidth}{!}{
    \begin{tabular}{p{2cm}ccccccccccc} \toprule
    & \multicolumn{2}{c}{mDETR}  & \multicolumn{2}{c}{MCN} & \multicolumn{2}{c}{Tanaka} & \multicolumn{2}{c}{Yu} &\multicolumn{2}{c}{ReCLIP} \\
    \cmidrule(lr){2-3} \cmidrule(lr){4-5} \cmidrule(lr){6-7} \cmidrule(lr){8-9} \cmidrule(lr){10-11}\\
     & Test A & Test B &  Test A & Test B & Test A & Test B & Test A & Test B & Test A & Test B\\
    \midrule 
    Schutz et al. & 34.8 & 26.4 &  \cellcolor{blue!25}{28.8} & \cellcolor{blue!25}{19.8} & 31.2 & 36 & 31.6 & 37.6  & 44.6 & 44.2\\
    Tanaka et al.  & 22.4 & 19.8 & 16.4 & 15.4 & \cellcolor{blue!25}{21.4} & \cellcolor{blue!25}{27.0} & 30.0 & 36.0 & 20.4 &  29\\
    Yu et al.  & 27.4 & 22.2  & 20.2 & 14.4  & \textbf{35.0} & \textbf{40.6} & \cellcolor{blue!25}{38.0} & \cellcolor{blue!25}{41.4} & 22.6 & 28.2 \\
    \midrule 
    \disclip{} & \textbf{36.2} & \textbf{30.8} & \textbf{24.4} & \textbf{15.6} & 27.6 & 32.0 & 30.4 & 32.2 & \cellcolor{blue!25}{66.2} & \cellcolor{blue!25}{68.6} \\
    \bottomrule
\end{tabular} }
\vspace{6pt}
\caption{\textbf{The effect of speaker-listener pairing.} Accuracy on RefClef dataset.} \label{tab:atificial-paring}
\end{table}

\section{Robustness to hyper-parameters} \label{sec:hpt}
\begin{figure}[h!]
\centering
    \includegraphics[width=0.7\textwidth]{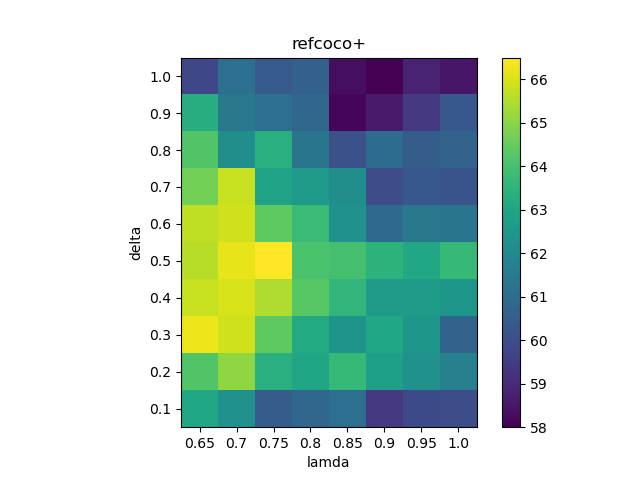}
    \caption{Sensitivity to Hyperparameters. Shown is the accuracy on RefCOCO+ as a function of various values of $\delta$ and $\lambda$. The color bar represents the Acc@0.5. Optimal values are obtained for $\lambda = 0.75$ -- assigning significant weight to distractor boxes, but smaller than half, and with $\delta = 0.5$ -- assigning equal weight to the local and global representation of the box in the image.}
    \label{fig:tuning}
\end{figure}

DisCLIP requires no training, but we tuned its hyperparameters $\delta$ and $\lambda$ on a subset (1000 random samples from 3805) of RefCOCO+ validation split, see Figure \ref{fig:tuning}. In all cases, we used the "natural" listener that is "paired" with the speaker in the sense that the listener was used either when training or evaluating the speaker in the original papers. 
 
\section{DisCLIP Objective}\label{sec:supp_loss}
The DisCLIP model consists of two branches: a language branch where a large language model (LM) generates a sequence, and a visual branch that guides the generation process towards a target image region in a visual-semantic space. The overall objective is defined as:
\begin{equation}
    v = \argmax_{v\in V^{(k)}} \Big \{ \mathcal{L}_{lang} + \beta \cdot \mathcal{L}_{DisCLIP} \; \Big \},
\end{equation}
where $v$ represents the next candidate token, $V^{(k)}$ denotes the set of top-$k$ predictions from the model's probability distribution, and $\beta$ is a hyper-parameter controlling the trade-off between language and vision. When $\beta=0$, the visual controls are disabled.

In the main text, we discuss the vision part $\mathcal{L}_{DisCLIP}$, which maximizes the similarity \cite{hessel2021clipscore} between the generated sequence and a specific \textit{region} in the image, while minimizing the similarity to a set of distractor regions.

The complete objective also includes two additional terms designed to ensure language fluency and consistency with the context tokens, referred to as $\mathcal{L}_{lang}$. These terms were defined in the optimization procedure proposed in \cite{su2022language}:

$$ \mathcal{L}_{lang} = (1\!-\!\alpha) \cdot \overbrace{p_\theta(v|x_{<t})}^{\text{model confidence}} - \alpha \cdot \overbrace{( max \{ s(h_v, h_{x_j}):1\leq j\leq t-1 \} )}^{\text{degeneration penalty}} $$

The last term was originally suggested in \cite{su2022contrastive} and addresses the degeneration problem in text generation by language models. This problem refers to the generation of dull and repetitive text at different levels (e.g., token-, phrase-, and sentence-level). The authors propose the use of contrastive learning to calibrate the representation space of the language model. $h_v$ represents the CLIP embedding of the sequence so far $x_{<t}$ and the current token $v$. We incorporate the degeneration penalty, along with the model confidence, into the score to guide the model towards likely outputs while avoiding the problem of model degeneration.

\section{Human Evaluation} \label{sec:supp_mturk}

\begin{figure}[ht]
    \centering
    \includegraphics[width=.6\textwidth, trim={0 2in 0in 0.5in}, clip]{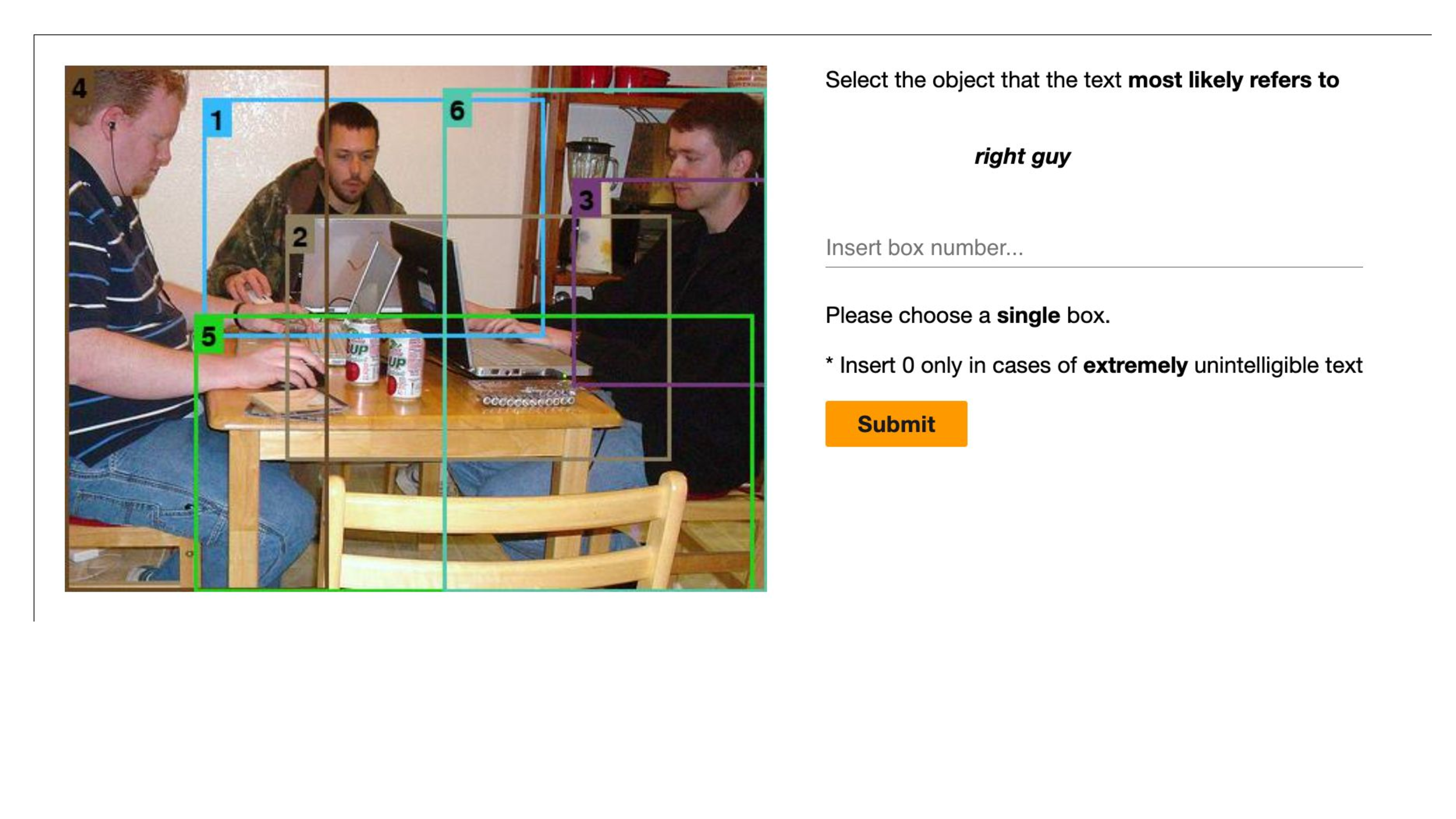} 
    \caption{\textbf{mTurk REC task.} Given a sentence, generated by our model or the baselines, we ask raters to select the box the text most likely refers to.}
    \label{fig:mturk_exp2}
\end{figure}

We utilized the Amazon Mechanical Turk (AMT) platform to conduct a direct comparison of the generated referring expressions (REs) in a REC task performed by human participants. In this task, participants were presented with multiple candidate boxes (n) and asked to select the one that best corresponds to the provided textual description. The textual descriptions were generated by our system as well as the baseline methods. The task layout is depicted in Figure~\ref{fig:mturk_exp2}. To ensure robust evaluation, we collected evaluations for 100 randomly selected samples from each of the three out-of-domain datasets. Each sample was independently evaluated by three distinct annotators. 

\subsection{Naturalness}\label{sec:sup_mturk_qual}
DisCLIP outperforms the baseline methods by generating more diverse and natural phrases, as demonstrated in the Flickr30k-Entities dataset. The wins and losses of our model compared to the baselines are presented in Table~\ref{tab:mturk_wins} and Table~\ref{tab:mturk_lose}, respectively.

\begin{table}[b] 
\centering
    \resizebox{0.75\textwidth}{!}{%
    \renewcommand*{\arraystretch}{1.5}
            \begin{tabular}{p{1in} |p{1in} |p{1in} |p{1in}} %
                & (a) & (b) & (c)  \\
                &  \includegraphics[width=1in, trim={0 2in 8in 2in}, clip]{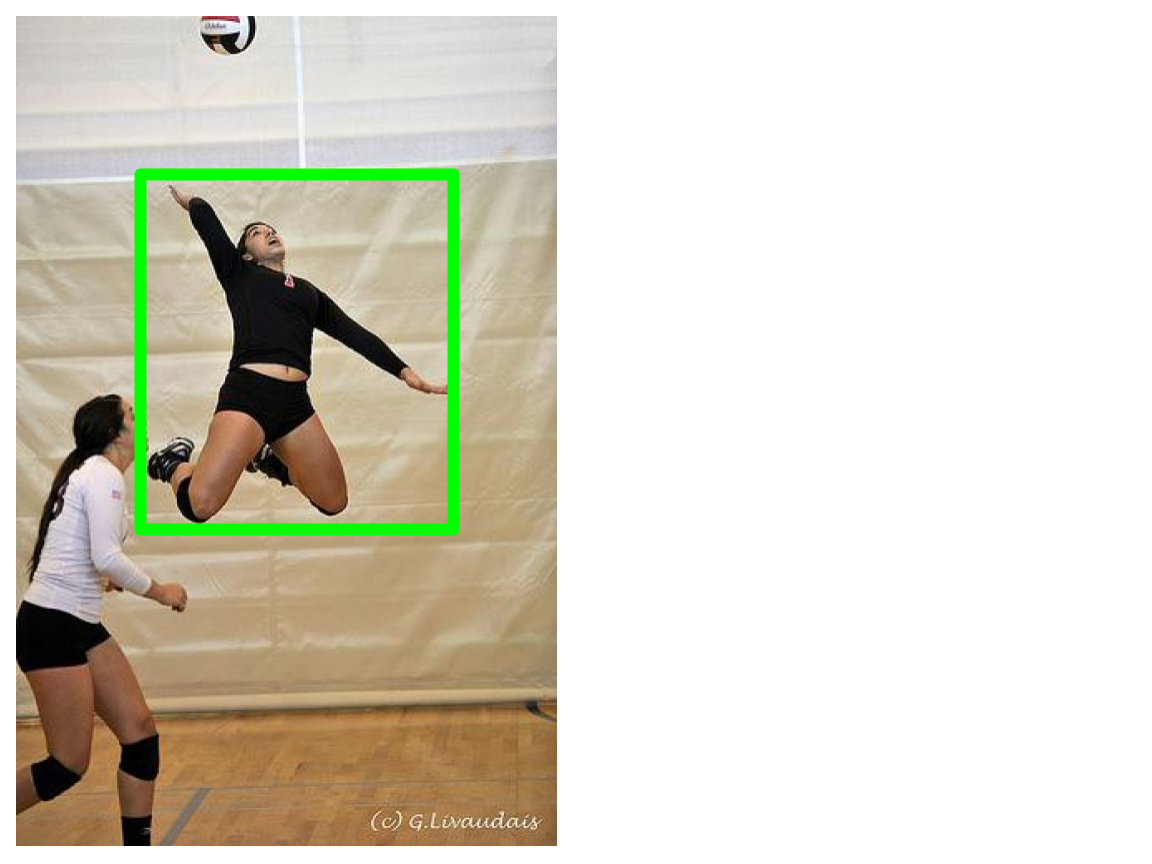} &  \includegraphics[width=1in, height=0.7in]{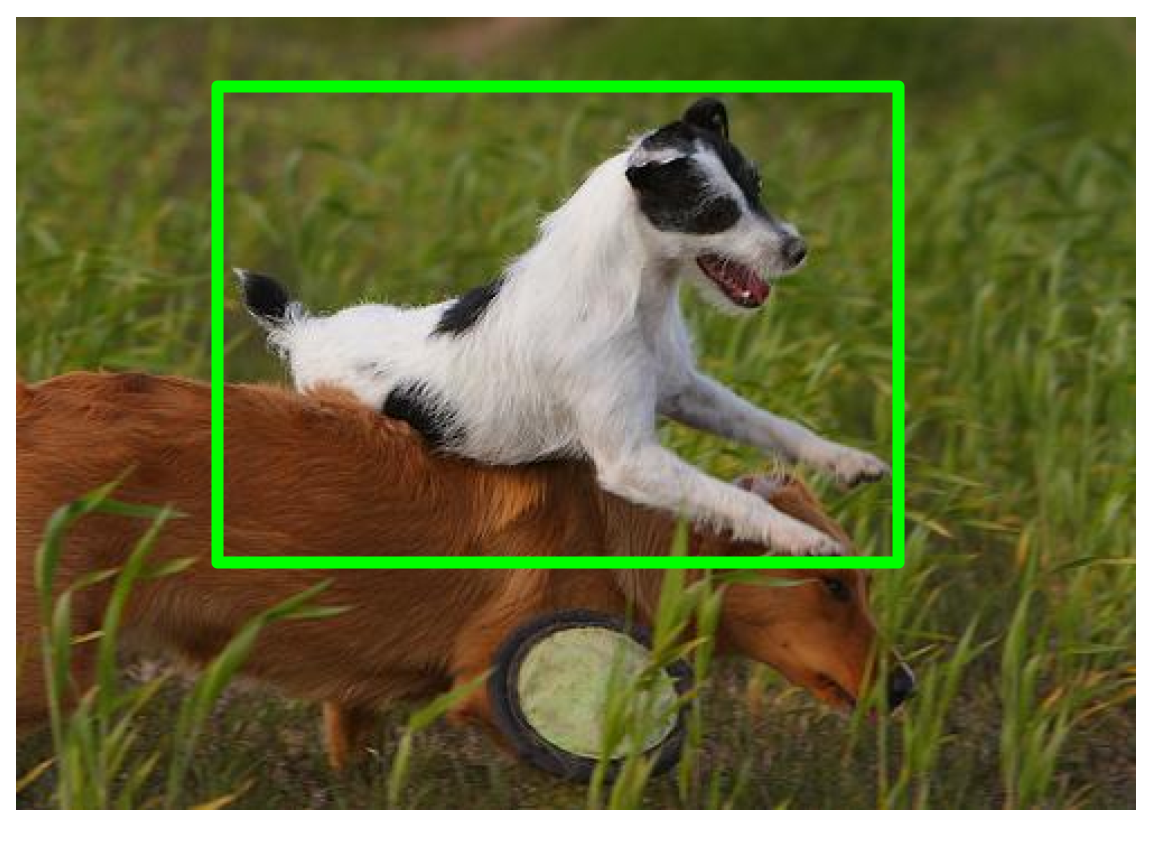} & \includegraphics[width=1in, height=0.7in]{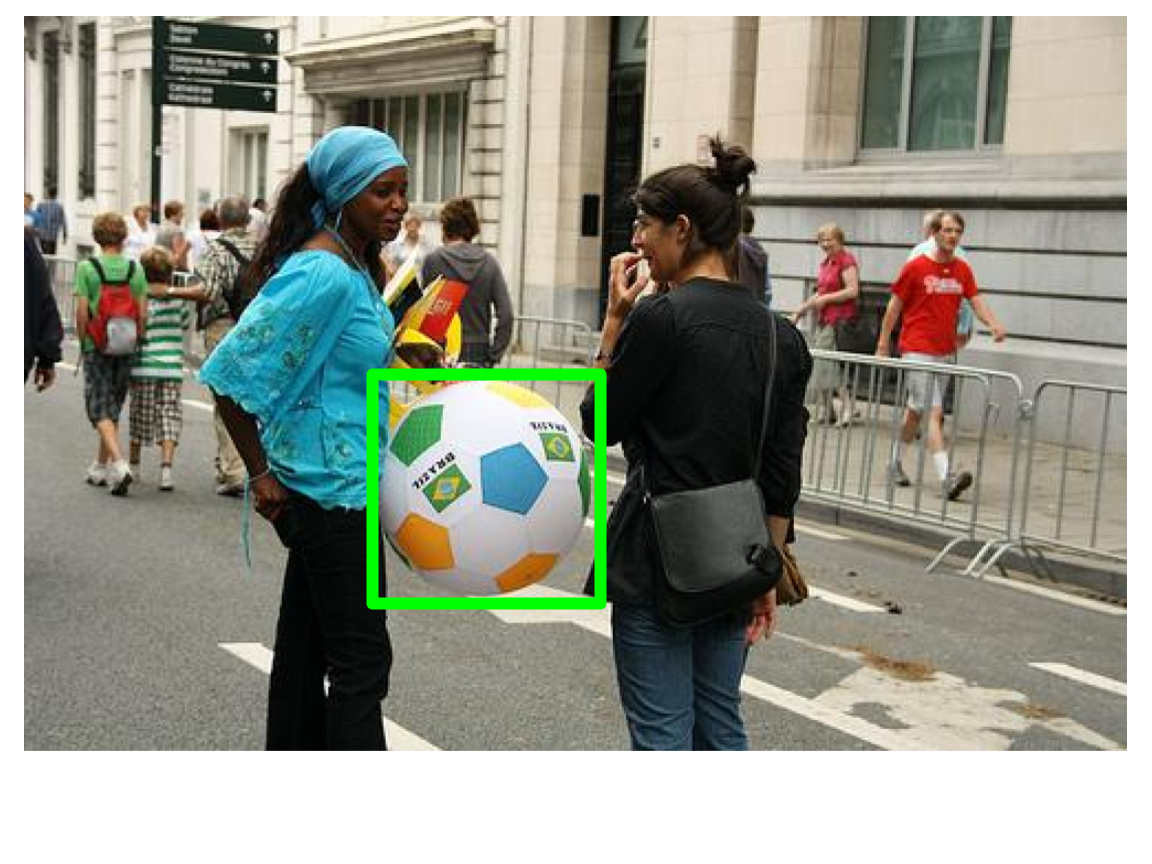} \\ \hline
                Schutz et al.\cite{schuz2021decoupling}  & Skateboarder  & black dog & number 1 meter \\ \hline
                Tanaka et al. \cite{tanaka2019generating}  & Woman & barber barber barber barber & barber barber barber barber barber \\ \hline
                Yu et al. \cite{yu2017joint}  & Woman & camper goo goo mix raspberries goo mix raspberries & curlys curlys curlys curlys \\ \hline
                \textsc{DisCLIP-HPT (Ours)}  & \textbf{Woman jumping} as part of volleyball swing using \textbf{black} object. & \textbf{White dog} riding horse during a dog fight. & The baseballs of teams one \textbf{yellow} \textbf{white} \textbf{green} \textbf{orange} \\ %
            \end{tabular}  } \vspace{6pt}
    \caption{\textbf{Win cases from Flickr30k Entities.} In all these examples, people successfully chose the target object given our caption, but failed to do so, given the captions produced by the baselines.} \label{tab:mturk_wins}
\end{table}

In Table~\ref{tab:mturk_wins}a, for example, two baselines produce accurate but non-discriminative captions such as "\textit{woman}". In contrast, our model specifies "\textit{woman jumping}" uniquely identifying the target object. Additionally, it provides context (e.g., "\textit{volleyball}") and information about attributes (e.g., "\textit{black}"). In Table~\ref{tab:mturk_wins}c, all models struggle. However, the raters found the descriptions of all other baselines unintelligible, except for ours. DisCLIP offers enough clues regarding visual attributes like "\textit{yellow}" "\textit{white}" "\textit{green}" and "\textit{ball}" aiding the raters in correctly identifying the colorful ball.

Table~\ref{tab:mturk_lose} presents examples where our method did not exhibit a clear advantage. In Table~\ref{tab:mturk_lose}a, individuals correctly identified the target object in all cases, but DisCLIP provided more information than necessary. Table~\ref{tab:mturk_lose}b illustrates a common issue where DisCLIP captures unwanted contextual information in cases of overlapping boxes. In the given example, the target object is the white shirt, but DisCLIP focuses on the woman and entirely omits the shirt. We anticipate that utilizing segmentation masks instead of boxes will help mitigate these problems.

\begin{table}[t] 
\centering
    \renewcommand*{\arraystretch}{1.5}
    \resizebox{0.75\textwidth}{!}{%
    \begin{tabular}{p{1in}|p{1in}|p{1in}|p{1in}} %
        & (a) & (b) & (c)  \\
        &  \includegraphics[width=1in, height=0.7in]{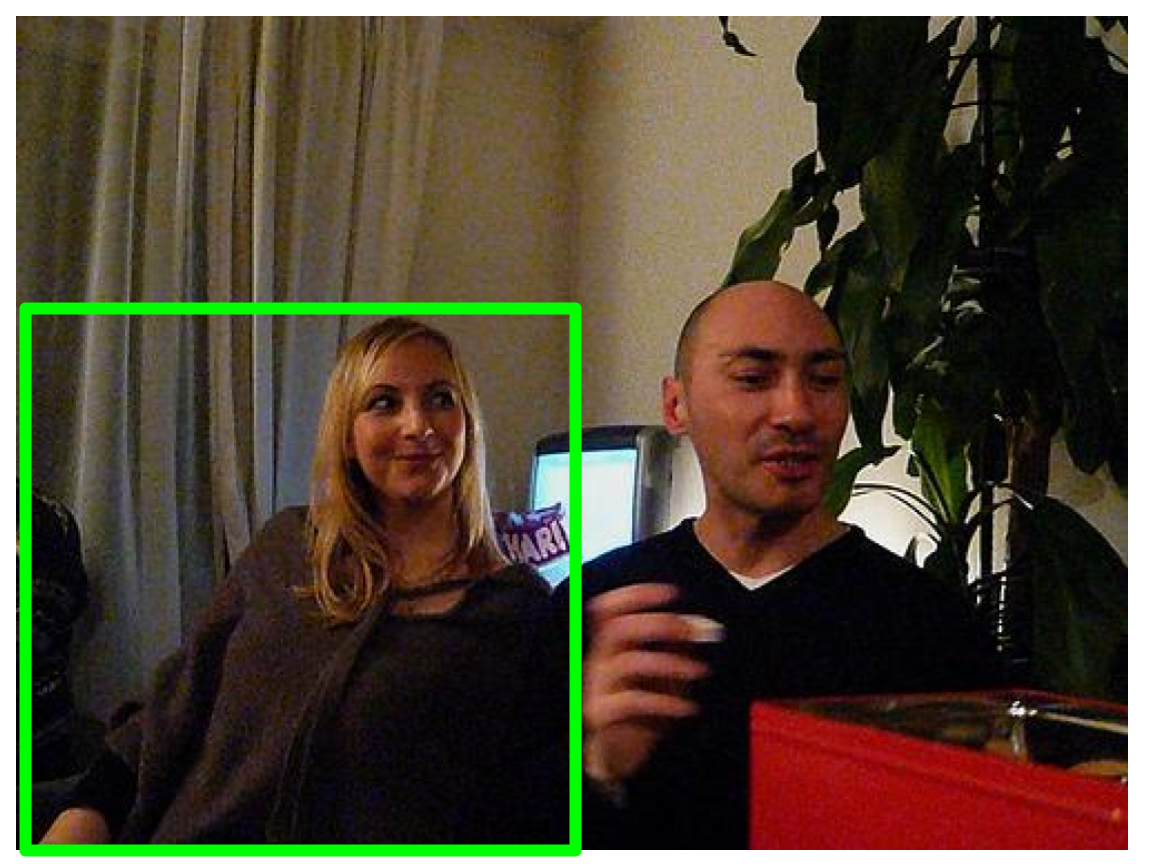} & \includegraphics[width=1in, height=0.7in]{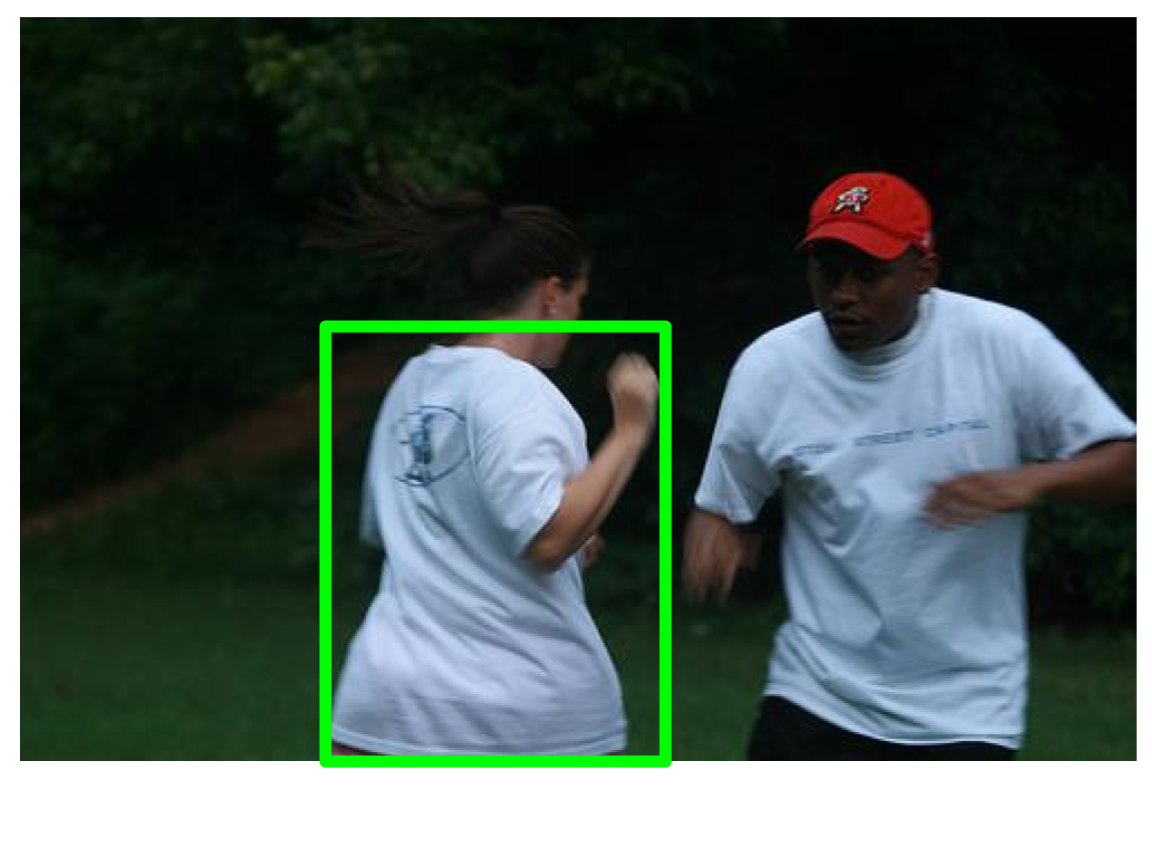}  & \includegraphics[width=1in, height=0.7in]{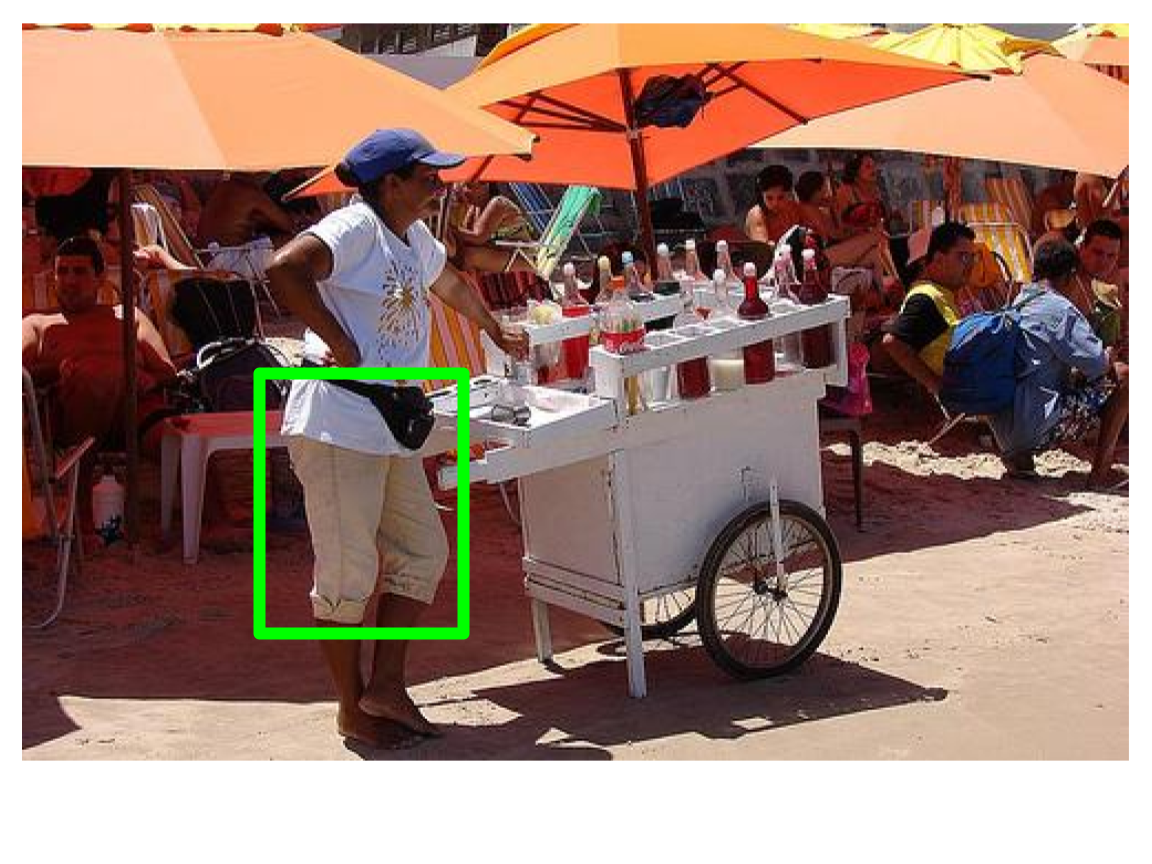} \\ \hline
        Schutz et al.\cite{schuz2021decoupling}  & younger woman & white shirt & white pants \\ \hline
        Tanaka et al. \cite{tanaka2019generating}  & woman& white shirt & barber barber barber barber barber \\ \hline
        Yu et al. \cite{yu2017joint}  & woman & white shirt & partial partial partial barely partial barely fingers curlys submerged  \\ \hline
        \textsc{DisCLIP-HPT (Ours)}  & Woman smiling posing with multiple pairs of black neck shirts & Woman standing eating & Person dressed shirt pants leg boots while holding phone
    \end{tabular} 
    } \vspace{6pt}
    \caption{\textbf{Lose cases from Flickr30k Entities.}} \label{tab:mturk_lose}
\end{table}

\subsection{Diversity}
Models based on open text generation naturally have a larger vocabulary compared to supervised methods, which are trained on a limited, predetermined set of categories. In the case of the Flickr30k-test dataset, the baselines achieve a maximum vocabulary of 519 words, while our models cover a much larger vocabulary of 4279 words, which is more than \textbf{eight times} the size.

\begin{figure}[hb]
    \centering
    \includegraphics[width=0.7\linewidth, trim={0 0.5in 0 0.3in}, clip]{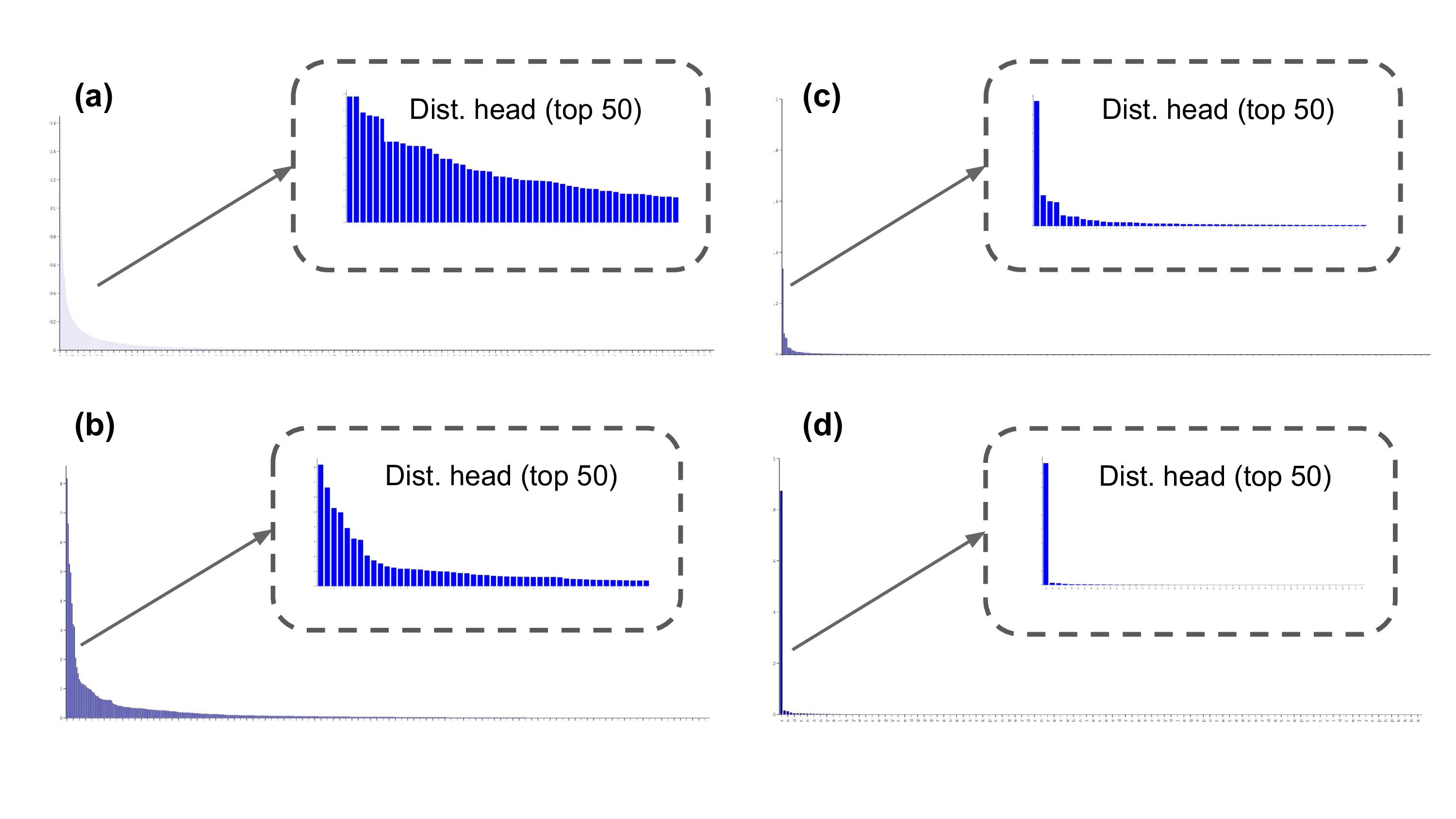} 
    \caption{Distribution of words in the generated referring expression in Flickr30k-Entities by the different models: (a) DisCLIP (b) Schutz (c) Yu (d) Tanaka }
    \label{fig:dist}
\end{figure}

\begin{table}[ht] 
\centering
    \renewcommand*{\arraystretch}{1.5}
    \resizebox{\textwidth}{!}{%
            \begin{tabular}{ll} \toprule
                    &  \textbf{Top 10 words} \\ \midrule
                Schutz et al.\cite{schuz2021decoupling}    & shirt (782), man (635), red (503), white (475), black (374), blue (306), woman (298), green (197), blurry (166) , dog (146)  \\ \hline
                Tanaka et al. \cite{tanaka2019generating}  & barber (25081), shirt (456), man (366), white (222), red (149), black (140), closest (136), blue (128), woman (110), barely (109) \\ \hline
                Yu et al. \cite{yu2017joint}  & curlys (9191), loops (2257), goo (1815), raisins (1745), almonds (775), seed (691), blurry (686), shirt (499), man (424), mix (398) \\ \hline
                \textsc{DisCLIP (Ours)}  & white (591), black (591), holding (516), red (502), person (498), woman (487), large (423), man (416), young (371), close (360) \\ \bottomrule
            \end{tabular} } \vspace{1pt}
    \caption{Most common words in the generated REs for Flickr30k-Entities test split. (overall 4601 Refs in 979 images)} \label{tab:top10}
\end{table}

\newpage
\section{Datasets for Referring Expressions Generation} \label{sec:supp_datasets}
Below, we provide some more details about the datasets that were used in the scope of this paper.

\noindent\textbf{(1) RefCOCO }\cite{kazemzadeh2014referitgame} contains 142,209 referring expressions for 50,000 objects in 19,994 images. 

\textbf{(2) RefCOCO+} \cite{kazemzadeh2014referitgame} contains 141,564 referring expressions for 49,856 objects in 19,992 images. This dataset focuses more on the appearance of objects. In both RefCOCO and RefCOCO+, Test A contains references to humans, and Test B references to other object types. 

\textbf{(3) RefCOCOg (Google RefExp)} \cite{mao2016generation} contains 85,474 referring expressions for 54,822 objects in 26,711 images and contains longer and more complex expressions. 

\textbf{(4) RefCLEF} (ReferIt) \cite{kazemzadeh-etal-2014-referitgame} 10,000 images for training/validation and 10,000 for test, with 59,976 references in the train/val set and 60,105 in the test set. RefCLEF dataset is larger and more varied and is curated specifically \textit{complex} photographs of real-world cluttered scenes. 

\textbf{(5) RefGTA} \cite{tanaka2019generating}, contain 6563 samples in train/val and 6504 in test. Synthetic images are from the Grand Theft Auto (GTA) videogame. All referring expressions in RefGTA correspond to people. The focus is on relations expressions, since for  a salient target, a brief description suffices. while less salient targets, require utilizing relationships with salient contexts around them to help tell their location. 

\textbf{(6) Flickr30k-Entities} \cite{plummer2015flickr30k}, provides a comprehensive ground-truth correspondence between regions in images and phrases in captions. It contains 244K coreference chains, with 275K corresponding bounding boxes. We excluded ``group'' references (e.g. \textit{People are outside waving flags}), resulting in 1966 images and 4597 references in the validation set and 4601 in the test.  

\begin{figure}[hb]
    \centering
    \includegraphics[width=0.9\linewidth, trim={0 0.7in 0 0.7in}, clip]{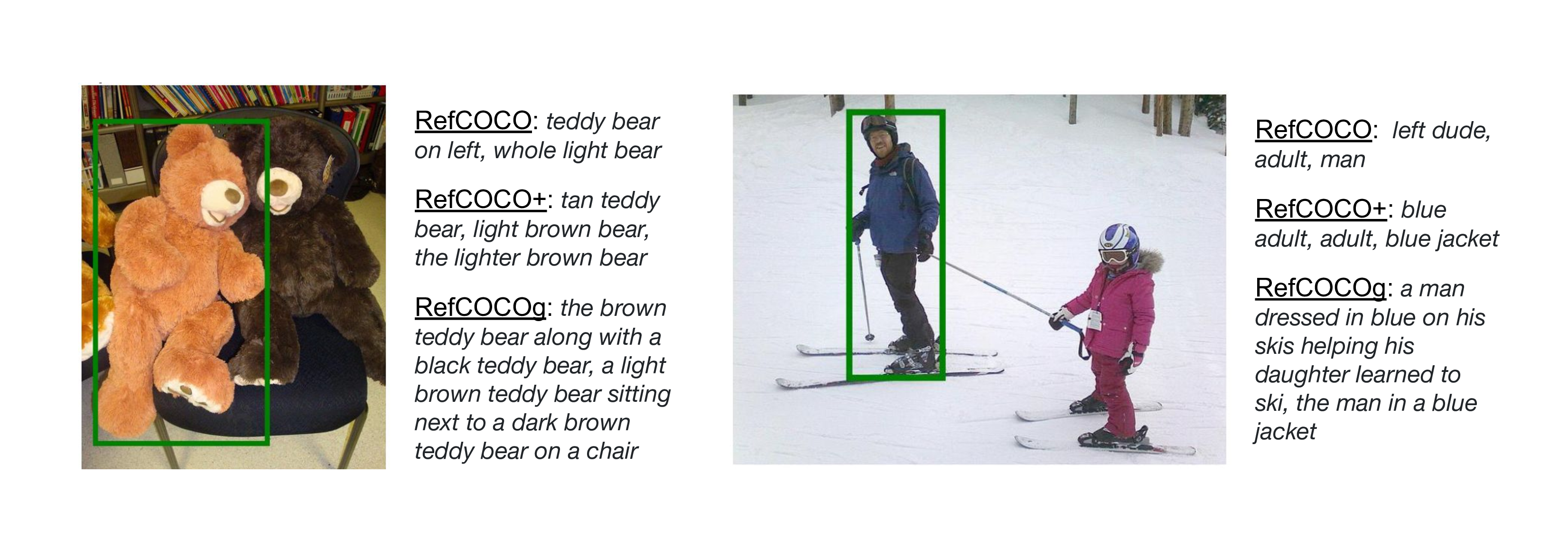} 
    \caption{ Images in RefCOCO/+/g are from MS-COCO dataset, but their textual annotations were designed to capture different types of referring expressions. RefCOCO is focused on spatial phrases, RefCOCO+ is attribute-based, and RefCOCOg provides long, rich and diverse text. }
    \label{fig:datasets}
\end{figure}

\end{document}